\documentclass[11pt]{article}

\usepackage[final]{acl}

\usepackage{times}
\usepackage{latexsym}
\usepackage{amsmath}
\usepackage{amssymb}
\usepackage{booktabs}
\usepackage{multirow}
\usepackage{subcaption}
\usepackage[table]{xcolor}
\usepackage[most]{tcolorbox}
\usepackage{xspace}
\usepackage{placeins}
\usepackage[ruled,vlined,linesnumbered]{algorithm2e}
\usepackage{xcolor}
\usepackage{enumitem}
\usepackage{fancyvrb}
\usepackage{fvextra}
\usepackage{tikz}
\usetikzlibrary{arrows.meta}

\SetCommentSty{bluecomment}
\SetKwComment{tcp}{$\triangleright$~}{}
\SetKwInput{KwData}{Input}
\SetKwInput{KwResult}{Output}
\SetAlgoNlRelativeSize{-1}
\DontPrintSemicolon

\usepackage[T1]{fontenc}
\usepackage[utf8]{inputenc}

\usepackage{microtype}

\usepackage{inconsolata}

\usepackage{graphicx}
\usepackage{booktabs}

\newcommand{\ours}{SafeRx-Agent\xspace}

\title{SafeRx-Agent: A Knowledge-Grounded Multi-Agent Framework for Safe and Explainable Medication Recommendation} 

\author{
\textbf{Xinyu Wang}\textsuperscript{1}\thanks{Equal contribution.},
\textbf{Hanwei Wu}\textsuperscript{2}\footnotemark[1],
\textbf{Zhenghan Tai}\textsuperscript{3}\footnotemark[1],
\textbf{Sicheng Lyu}\textsuperscript{1},  
\textbf{Qincheng Lu}\textsuperscript{1}, \\
\textbf{Ziyu Zhao}\textsuperscript{1}, 
\textbf{Jijun Chi}\textsuperscript{3}, 
\textbf{Jingrui Tian}\textsuperscript{1}, 
\textbf{Xiao-Wen Chang}\textsuperscript{1}, 
\textbf{Ziyang Song}\textsuperscript{4}\thanks{Corresponding author.}
\\[0.5em]
\textsuperscript{1}McGill University 
\textsuperscript{2}McMaster University 
\textsuperscript{3}University of Toronto
\textsuperscript{4}Ohio University
}

\begin{document}
\maketitle

\begin{abstract}
Medication recommendation predicts medications for patient visits, but existing methods still face two key challenges. At the model level, traditional drug recommendation methods only predict structured drug codes with limited evidence grounding, while LLM agents can use richer clinical context but may lack safety verification and traceability. At the task level, existing benchmarks often use broad medication categories, which ignore subgroup-level safety differences and can lead to risk overestimation. We introduce the first fine-grained medication recommendation setting based on fourth-level ATC code generation. We propose \textbf{Safe Prescription Agent (SafeRx-Agent)}, a knowledge-grounded multi-agent framework that uses patient context, external clinical knowledge, and safety verification to recommend traceable medication sets. Experimental results on MIMIC-III and MIMIC-IV datasets show that SafeRx-Agent improves fine-grained medication prediction accuracy while controlling drug interactions, contraindications, and medication set size.
\end{abstract}

\section{Introduction}

Medication recommendation from electronic health records (EHRs) is a high-stakes clinical natural language processing (NLP) task~\cite{xu2022survey}. Given a patient's longitudinal clinical context, including prior visits, diagnoses, procedures, and medication history, the model predicts the medications for the current encounter. This task is challenging because ICU histories are sparse, visits involve multiple active conditions, and prescribing depends on both acute illness and treatment continuity~\cite{Shang_Xiao_Ma_Li_Sun_2019}.

Deep learning methods have advanced medication prediction by modeling visit sequences and drug co-occurrence~\cite{Ali2023-mh}, but they operate over structured EHR codes and cannot utilize patient-specific textual context or external medical evidence at inference time~\cite{Shang_Xiao_Ma_Li_Sun_2019,ijcai2021p514}. Large language models (LLMs) can process clinical text and generate medication sets, and multi-agent systems further decompose complex clinical reasoning into coordinated steps with tool use~\cite{liu2025largelanguagemodeldistilling, fan2026finegrainedlistwisealignmentgenerative, li2024mmedagentlearningusemedical}. However, unconstrained LLM agents remain unreliable for clinical decision support: prior work reports hallucinations, guideline misalignment, and unsafe recommendations~\cite{hager2024evaluation,asgari2025framework,farrag2026evaluating}. Safe medication recommendation therefore requires a knowledge-grounded agentic framework with explicit evidence and safety verification.

% However, most of these methods operate over structured EHR codes and provide limited support for using patient-specific context and external medical evidence to guide medication generation. In contrast, large language models (LLMs) can process textual clinical context, longitudinal patient profiles and generate medication sets from patient-specific information. Multi-agent systems extend these capabilities by adding tool use and decomposing complex tasks into coordinated reasoning steps. Despite these capabilities, unconstrained LLM agents remain unreliable for clinical decision support. Such agents may return plausible medication sets without reliable evidence, safety checks, or verifiable rationales. Prior evaluations have reported hallucinations, guideline misalignment, and inaccurate or unsafe treatment recommendations from medical LLMs~\cite{Hager2024-tr,Asgari2025-my,Farrag2026-rf}. Medication recommendation therefore requires a knowledge-grounded agentic framework for safe and verifiable medication generation.

%  assigning patient understanding, candidate generation, safety checking, and reporting to coordinated roles, often with tool use.   are not inherently grounded in patient-specific evidence, medical knowledge, safety constraints, or traceable justification. They may produce invalid ATC codes, recommend medications unsupported by the patient's conditions, miss drug--drug interactions or disease-conditioned contraindications, or provide rationales that clinicians cannot verify.   that  medical evidence, safety verification, and traceable reporting.

Medications are standardized with the Anatomical Therapeutic Chemical (ATC) taxonomy~\cite{who2026atc}, which organizes drugs into five hierarchical levels. Most benchmarks predict at the third level of the ATC taxonomy, denoted ATC-L3, which merges medication subgroups that can differ in clinical use and safety profile~\cite{Ali2023-mh}. This distorts safety evaluation: a drug interaction may apply to one fine-grained subgroup but not another under the same ATC-L3 parent, causing ATC-L3 evaluation to overestimate risk. Accurate safety measurement therefore requires predicting medication codes at a finer granularity.

% Medication codes are commonly defined with the Anatomical Therapeutic Chemical (ATC) taxonomy~\cite{atc}, which organizes drugs into five levels from broad anatomical groups to fine-grained chemical substances. Most medication recommendation benchmarks use coarse drug categories, often third-level ATC codes (ATC-L3)~\cite{Ali2023-mh}, because this reduces the prediction space. This simplification merges medication subgroups that may differ in clinical use and safety profile, which can distort safety measurement. A drug interaction or disease-conditioned contraindication may apply to one fine-grained medication subgroup but not to another under the same ATC-L3 parent code. ATC-L3 safety evaluation can therefore overestimate risk, because a parent category is flagged as risky when any of its fine-grained subgroups has a recorded risk. Accurate and safe medication recommendation therefore requires fine-grained prediction that can distinguish clinically related medication subgroups and measure safety risks more precisely.

We propose \textbf{Safe Prescription Agent (\ours)}, a knowledge-grounded multi-agent framework for safe and explainable fine-grained medication recommendation. \ours routes each patient case to specialty-aware medication agents that generate fine-grained ATC-L4 medication candidates grounded in patient context, ICD and ATC taxonomies, and medication indication evidence. A safety-aware critic-revision loop then checks candidates against drug-drug interaction (DDI) and contraindication resources, revises unsafe predictions, and produces a traceable report. We evaluate \ours on MIMIC-III~\citep{mimiciii} and MIMIC-IV~\citep{mimiciv} using both medication prediction accuracy and safety metrics. \ours outperforms traditional deep learning, LLM, and agentic baselines in prediction accuracy while reducing DDI and contraindication rates through explicit safety verification.
Our key contributions are:
\begin{itemize}[leftmargin=1em, itemsep=0pt, topsep=2pt, parsep=0pt]
    \item We introduce the first fine-grained medication recommendation setting for predicting ATC-L4 code sets from EHRs, moving beyond the coarse ATC-L3 setting used in prior benchmarks.
    \item We propose \ours, a medication recommendation multi-agent framework that combines specialty-aware generation, evidence grounding, safety-aware revision, and traceable reporting in a unified workflow.
    \item We introduce a knowledge-grounded safety verifier that detects the risks of drug interactions and contraindications, revises unsafe candidates, and produces traceable medication reports.
    \item We evaluate \ours on two real-world EHR datasets, showing improved fine-grained prediction accuracy, lower safety risks, and predicted set sizes closer to the ground truth.
\end{itemize}

\section{Related Work}

\paragraph{Supervised Medication Recommendation.}
Supervised methods cast medication recommendation as multi-label
prediction over structured EHR codes, with
GAMENet~\citep{Shang_Xiao_Ma_Li_Sun_2019} introducing graph-augmented
memory and DDI-aware decoding, and follow-ups adding molecular
structure~\citep{ijcai2021p514, molerec}, copy-generate
decoding~\citep{cognet}, and rare-drug or cold-start
training~\citep{raremed, drugdoctor}. These models operate on coarse ATC-L3 vocabularies and encode safety implicitly in the loss, but fail to perform fine-grained medication prediction.

\paragraph{LLMs for Medication Recommendation.}
Direct prompting of general or medical
LLMs~\citep{ultramedical, med42v2, OpenBioLLMs,
chen2024huatuogpto1medicalcomplexreasoning,
garciagasulla2025aloefamilyrecipeopen} leverages richer textual context
than structured-code models, but lacks built-in safety
verification~\citep{hager2024evaluation, asgari2025framework,
farrag2026evaluating}. Fine-tuning approaches such as
LAMO~\citep{zhao2025finegrainedalignmentlargelanguage},
FLAME~\citep{fan2026finegrainedlistwisealignmentgenerative}, and
LEADER~\citep{liu2025largelanguagemodeldistilling} address safety via
losses, rewards, or distillation, but require task-specific training and
remain tied to fixed backbones.

\paragraph{Multi-Agent Frameworks for Clinical Decision Support.}
Multi-agent LLM systems decompose clinical reasoning across coordinated
roles for medical QA~\citep{medagents, mdagents} and rare-disease
diagnosis and treatment~\citep{rareagents}. However, existing agent-based generation frameworks generally do not support fine-grained medication prediction with resource-grounded verification of multiple safety risks, including DDIs and diagnosis-conditioned contraindications.

\section{Problem Formulation and Knowledge Resources}
\label{sec:task}

\subsection{Problem Formulation}
Let a patient record be a temporally ordered sequence of ICU visits $(v_1,\ldots,v_T)$, where each visit
$v_t=(\mathcal{D}_t,\mathcal{P}_t,\mathcal{M}_t)$ contains diagnoses, procedures, and medications. Diagnoses are represented by ICD-CM codes, procedures by ICD-PCS codes, and medications by ATC-L4 codes.  Given the past visits and the current diagnoses and procedures, the task is to predict the medications prescribed at the current visit. The input for visit $T$ is
\[
X_T=\big(\{(\mathcal{D}_t,\mathcal{P}_t,\mathcal{M}_t)\}_{t<T},
\mathcal{D}_T,\mathcal{P}_T\big).
\]
The ground-truth medication set is $\mathcal{M}_T \subseteq \mathcal{V}_{\mathrm{med}}$,  and the model outputs a predicted set $\hat{\mathcal{M}}_T \subseteq \mathcal{V}_{\mathrm{med}}$,  where $\mathcal{V}_{\mathrm{med}}$ is the ATC-L4 medication vocabulary. We evaluate prediction quality by comparing
$\hat{\mathcal{M}}_T$ with $\mathcal{M}_T$, and evaluate safety by measuring DDIs and contraindications in $\hat{\mathcal{M}}_T$.

\subsection{Clinical Knowledge Resources}
SafeRx-Agent uses external clinical knowledge for evidence-grounded medication generation and safety-aware revision. We organize these resources around three functions:
\begin{itemize}
    \setlength\itemsep{0pt}
    \item \textbf{Standardization:} diagnosis and medication taxonomies align EHR concepts with the ATC-L4 medication vocabulary.
    \item \textbf{Grounding:} indication evidence links patient conditions to clinically relevant medications.
    \item \textbf{Safety checking:} drug interaction and contraindication resources identify co-prescription and disease-conditioned risks.
\end{itemize}
% These resources are especially important in the ATC-L4 setting, where
% fine-grained medication subgroups may differ in clinical use and safety profile. 
Table~\ref{tab:knowledge_resources} summarizes the resources used. Details about preprocessing, identifier mapping, and matrix construction are provided in Appendix~\ref{app:resources}.

\begin{table*}[t]
    \centering
    \scriptsize
    \setlength{\tabcolsep}{0.5pt}
    \caption{\textbf{Clinical knowledge resources used in this work.}
    All medication-side resources are mapped to the ATC-L4 vocabulary. Contra.\ = Contraindication.}
    \label{tab:knowledge_resources}
    \begin{tabular}{@{}p{0.30\textwidth} p{0.42\textwidth} p{0.28\textwidth}@{}}
    \toprule
    \textbf{Resource} & \textbf{Derived representation} & \textbf{Used in SafeRx-Agent} \\
    \midrule
    ICD/CCS taxonomy~\cite{icd}
    & Diagnosis codes grouped into clinical chapters and subcategories
    & Expert panel \& routing (Section~\ref{sec:expert_panel} \&~\ref{sec:generation}) \\
    \midrule
    ATC taxonomy~\cite{atc}
    & Defines ATC-L4 prediction vocabulary $\mathcal{V}_{\mathrm{med}}$
    & Throughout Section~\ref{sec:method} \\
    \midrule
    MEDI~\cite{medi}
    & Diagnosis--medication indication relation $R \subseteq \mathcal{V}_{\mathrm{diag}} \times \mathcal{V}_{\mathrm{med}}$
    & Generation \& critique (Section~\ref{sec:generation}) \\
    \midrule
    TWOSIDES~\cite{Tatonetti2012-af}
    & DDI matrices ${\mathbf{M}_{\mathrm{DDI}}^{\mathrm{bin}},\mathbf{M}_{\mathrm{DDI}}^{\mathrm{freq}}\!\in\!\mathbb{R}^{|\mathcal{V}_{\mathrm{med}}|^{2}}}$
    & Safety verification (Section~\ref{sec:safety}) \\
    \midrule
    openFDA drug labels~\cite{Kass-Hout2016-vx}
    & Contra. matrices ${\mathbf{M}_{\mathrm{Contra}}^{\mathrm{bin}},\mathbf{M}_{\mathrm{Contra}}^{\mathrm{freq}}\!\in\!\mathbb{R}^{|\mathcal{V}_{\mathrm{med}}|\times|\mathcal{V}_{\mathrm{diag}}|}}$
    & Safety verification (Section~\ref{sec:safety}) \\
    \bottomrule
    \end{tabular}
\end{table*}

\section{SafeRx-Agent}
\label{sec:method}

\subsection{Overview}
\label{sec:method_overview}

\begin{figure*}[ht]
    \centering
    \includegraphics[width=0.9\linewidth]{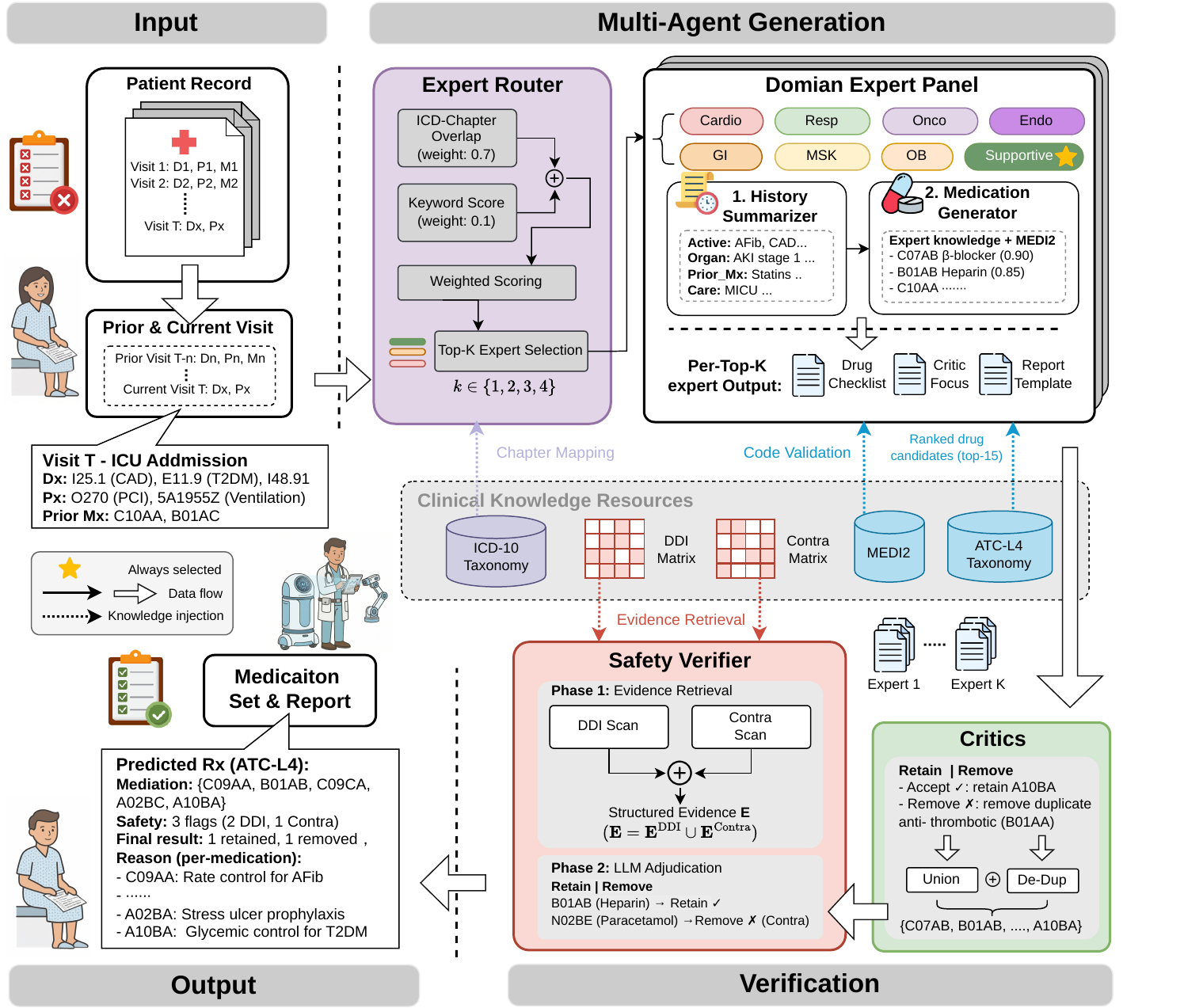}
    \caption{\textbf{Overview of SafeRx-Agent.} A patient record is routed via weighted ICD-chapter and keyword scoring to a sparse subset of specialty experts plus an always-on supportive-care expert. Each activated expert summarizes the patient record from its specialty scope and generates ATC-L4 medication candidates grounded in MEDI and the ATC taxonomy. A global Critique then judges and reconciles the expert proposals under the full patient context to produce a candidate set. A two-phase Safety Verifier retrieves DDI and contraindication evidence ($\mathbf{M}_{\text{DDI}}$, $\mathbf{M}_{\text{Contra}}$) and adjudicates each flag in context, yielding a traceable ATC-L4 prescription with per-medication rationales.}
    \label{fig:workflow}
    \vspace{-0.4cm}
\end{figure*}

As shown in Figure~\ref{fig:workflow}, SafeRx-Agent predicts the medication set $\hat{\mathcal{M}}_T \subseteq \mathcal{V}_{\mathrm{med}}$ for the target visit from the longitudinal patient record $X_T$. SafeRx-Agent uses the clinical knowledge resources in Table~\ref{tab:knowledge_resources} to ground medication generation and verify safety risks. SafeRx-Agent uses a multi-agent design because ATC codes span heterogeneous therapeutic domains, and clinically plausible candidates may still be unsafe due to DDIs or contraindications. Algorithm~\ref{alg:saferx} summarizes the workflow, and the following subsections define each operator. The prompt templates used by the LLM-based operators are provided in Appendix~\ref{app:prompts}, Figures~\ref{fig:prompt_summarizer}--\ref{fig:prompt_safety_verifier}.

\begin{algorithm}[t]
\small
\RestyleAlgo{boxruled}
\caption{SafeRx-Agent Inference}
\label{alg:saferx}

\KwData{Patient record $X_T$; expert panel $\mathcal{E}$; indication
resource $R$; safety matrices $\mathbf{M}_{\mathrm{DDI}},
\mathbf{M}_{\mathrm{Contra}}$.}
\KwResult{Predicted medication set $\hat{\mathcal{M}}_T$.}

\BlankLine
\tcp{Multi-agent generation}
$A \leftarrow \textsc{Route}(X_T, \mathcal{E}),\ A \subseteq \mathcal{E}$
\nllabel{line:route}\;
\ForEach{$E_i \in A$}{
    $(s_i, \rho_i) \leftarrow \textsc{Summarize}(E_i, X_T)$
    \nllabel{line:summarize}\;
    $\mathcal{M}_i \leftarrow \textsc{Generate}(E_i, s_i, \rho_i, X_T, R)$
    \nllabel{line:generate}\;
}
$\tilde{\mathcal{M}} \leftarrow
\textsc{Critique}(\{\mathcal{M}_i, s_i, \rho_i\}_{i \in A}, X_T, R)$
\nllabel{line:critique}\;

\BlankLine
\tcp{Safety verification}
$\mathcal{F} \leftarrow \textsc{FindFlags}(\tilde{\mathcal{M}}, X_T,
\mathbf{M}_{\mathrm{DDI}}, \mathbf{M}_{\mathrm{Contra}})$
\nllabel{line:flags}\;
$\hat{\mathcal{M}}_T \leftarrow \tilde{\mathcal{M}}$\;
\ForEach{$f \in \mathcal{F}$}{
    \If{$\textsc{Verify}(f, \hat{\mathcal{M}}_T, X_T, R) = \textsc{Rem}$}{
        $\hat{\mathcal{M}}_T \leftarrow \hat{\mathcal{M}}_T \setminus \{m_f\}$
        \nllabel{line:remove}\;
    }
}
\Return{$\hat{\mathcal{M}}_T$}
\nllabel{line:return}\;
\end{algorithm}

\subsection{Expert Panel}
\label{sec:expert_panel}

We construct a reusable expert panel
$\mathcal{E}=\{E_1,\ldots,E_K\}\cup\{E_{\mathrm{univ}}\}$ before inference. The panel contains $K$ specialty experts and one always-on universal supportive-care expert $E_{\mathrm{univ}}$ for common inpatient medications.
We derive the $K$ specialty experts empirically by clustering patient embeddings and mapping each cluster to an ICD chapter-level scope. Appendix~\ref{app:expert_construction} provides implementation details.

\subsection{Multi-Agent Generation}
\label{sec:generation}
 
\noindent\textbf{\textsc{Route}: sparse specialty activation.}
Invoking the full expert panel for every patient wastes computation and dilutes specialty-specific signal with off-topic perspectives, so $\textsc{Route}$ activates only relevant experts by matching the patient's current and historical ICD codes in $X_T$ against each expert's scope, with $E_{\mathrm{univ}}$ always included:
\begin{itemize}
    \setlength\itemsep{0pt}
    \item $A \leftarrow \textsc{Route}(X_T, \mathcal{E})$, $A \subseteq \mathcal{E}$.
\end{itemize}
Routing is sparse, so the number of expert calls scales with case complexity rather than the full panel size $|\mathcal{E}|$.
\newline
\newline
\noindent\textbf{\textsc{Summarize}: expert-specific patient summarization.}
The raw record $X_T$ remains long and heterogeneous, with each specialty's relevant evidence scattered among entries that fall outside its concern. $\textsc{Summarize}$ therefore extracts and structures the specialty-relevant portions of $X_T$ into compact artifacts that surface what each expert needs for prescribing. For each activated expert $E_i$, $\textsc{Summarize}$ converts the
multi-visit record $X_T$ into expert-specific artifacts (Fig.~\ref{fig:prompt_summarizer}):
\begin{itemize}
    \setlength\itemsep{0pt}
    \item \textbf{Typed summary} $s_i$: a structured summary of medication-related evidence, including active problems, acute organ dysfunction, procedures, care context, medication-relevant risks, and prior prescriptions.
    \item \textbf{Visit rationales} $\rho_i$: structured explanations for prior visits that link diagnoses, procedures, and prescribed drugs, helping distinguish continued medications from transient treatments.
\end{itemize}

\noindent\textbf{\textsc{Generate}: per-expert medication proposal.}
With a focused view of the patient, each expert is now positioned to move from understanding to action and propose medications grounded in its own scope. For each activated expert $E_i$,  $\textsc{Generate}$ proposes medications from the typed summary $s_i$, visit rationales $\rho_i$, and diagnosis-linked candidates retrieved from the MEDI resource $R$ (Row 3, Table~\ref{tab:knowledge_resources}). It also consults the raw record $X_T$ for details missing from $s_i$ or $\rho_i$:
\begin{itemize}
    \setlength\itemsep{0pt}
    \item $\mathcal{M}_i \leftarrow \textsc{Generate}(E_i, s_i, \rho_i, X_T, R)$.
\end{itemize}

\noindent\textbf{\textsc{Critique}: global critique and reconciliation.}
Individual experts provide focused but partial prescribing views. Their
proposals may overlap, omit cross-specialty context, or include medications
that are locally plausible but weakly supported by the full patient record.
$\textsc{Critique}$ reviews all expert proposals together with the expert-specific summaries $s_i$, visit rationales $\rho_i$, indication evidence $R$, and the full record $X_T$, and
then merges duplicate candidates and removes medications that lack sufficient patient-level support:
\begin{itemize}
    \setlength\itemsep{0pt}
    \item $\tilde{\mathcal{M}} \leftarrow
    \textsc{Critique}(\{\mathcal{M}_i, s_i, \rho_i\}_{i \in A}, X_T, R)$.
\end{itemize}

\subsection{Safety Verification}
\label{sec:safety}

SafeRx-Agent verifies each generated candidate set because clinically plausible medications may still create co-prescription or disease-conditioned safety risks. Safety verification takes the candidate set $\tilde{\mathcal{M}}$ as input and does not generate new medications. It uses the DDI and contraindication matrices, $\mathbf{M}_{\mathrm{DDI}}$ and $\mathbf{M}_{\mathrm{Contra}}$, derived from the safety resources in Table~\ref{tab:knowledge_resources}, rows 4 and 5. The verifier first retrieves safety evidence and then uses patient context to decide an action for each flagged item.

\paragraph{\textsc{FindFlags}: matrix-based evidence collection.}
$\textsc{FindFlags}$ scans $\tilde{\mathcal{M}}$ against
$\mathbf{M}_{\mathrm{DDI}}$ and $\mathbf{M}_{\mathrm{Contra}}$ to collect a set of flagged items $\mathcal{F}$. Each flag is represented as $f = (m_f, \mathrm{rel}_f, \mathrm{evid}_f)$, where $m_f$ is the flagged medication, $\mathrm{rel}_f \in \{\textsc{DDI}, \textsc{Contra}\}$ is the triggering safety relation, and $\mathrm{evid}_f$ contains case-level evidence such as DDI degree and prior-prescription status.

\paragraph{\textsc{Verify}: case-specific adjudication.}
For each flagged item, $\textsc{Verify}$ outputs an action $a_f \in \{\textsc{Ret}, \textsc{Rem}\}$ based on the flag $f$, the patient context $X_T$, and the indication resource $R$ (Fig.~\ref{fig:prompt_safety_verifier}). \textsc{Ret} retains a flagged medication when the patient context supports its use, such as a previously tolerated DDI pair. \textsc{Rem} removes a medication when the risk lacks patient-specific justification, such as a contraindication against an active diagnosis without prior exposure. 
This design keeps evidence-supported medication generation and removes candidates with unsupported safety risks.

\section{Experiment Design}
\label{sec:experiments}

\subsection{Datasets and Preprocessing}
\label{sec:datasets}
We evaluate SafeRx-Agent on MIMIC-III and MIMIC-IV. Diagnoses and
procedures are represented by ICD codes with textual descriptions;
medications are normalized to ATC-L4 codes with within-visit duplicates
removed. The prediction vocabulary $\mathcal{V}_{\mathrm{med}}$ is the
closed set of retained ATC-L4 codes; out-of-vocabulary predictions are
discarded at evaluation time. We use patient-level splits with no leakage.
For MIMIC-IV, we hold out 100 cases for prompt development and evaluate
on the remaining 1{,}586; for MIMIC-III, we use the full 901-patient
cohort with prompts transferred from MIMIC-IV without re-tuning.
Supervised baselines use the same splits. Dataset statistics are
summarized in Table~\ref{tab:data_stats} (Appendix~\ref{app:data}).

\subsection{Baselines}
\label{sec:baselines}
We compare \ours{} against four groups of baselines, isolating the effects
of task-specific training, medical-domain adaptation, general LLM
capability, and agent workflow design.
(1) \textbf{Traditional deep learning}: GAMENet~\citep{Shang_Xiao_Ma_Li_Sun_2019},
SafeDrug~\citep{ijcai2021p514}, DrugDoctor~\cite{drugdoctor}, and
KEHGCN~\citep{Zhang2026-li}.
(2) \textbf{Medical LLMs}: UltraMedical-70B~\cite{ultramedical},
Med42-v2-70B~\cite{med42v2}, OpenBioLLM-70B~\cite{OpenBioLLMs},
Llama3.1-Aloe-Beta-70B~\cite{garciagasulla2025aloefamilyrecipeopen}, and
HuatuoGPT-o1~\cite{chen2024huatuogpto1medicalcomplexreasoning}.
(3) \textbf{General LLMs}: GPT-5.2, Claude Sonnet-4.6,
Qwen3-32B~\cite{qwen3}, and Gemma3-27B-IT~\cite{gemma3}.
(4) \textbf{Agent-based baselines}: RareAgents~\cite{rareagents} adapted for ATC-L4 code generation, and General Agent with a single ICU medication expert.

\subsection{Implementation Details}
\label{sec:implementation}

Supervised baselines are trained on the patient-level splits with an output head over the ATC-L4 vocabulary and hyperparameters tuned on the validation set. All LLM-based methods share the same serialized EHR input and closed ATC-L4 vocabulary. The general-agent baseline replaces \ours{}'s routed specialty experts with one generalist agent, and RareAgents is adapted to ATC-L4 generation. Full baseline implementations, prompts, and inference settings are in Appendix~\ref{app:implementation_details}.

\subsection{Evaluation Metrics}
\label{sec:metrics}

We evaluate each method along two axes: medication-set prediction quality and
safety. For prediction quality, we report Jaccard, precision, recall, F1, and
the average number of predicted ATC-L4 codes. For safety, we report
GT-normalized DDI rates (DDI-B and DDI-W) and contraindication rates (Contra-B and Contra-W), where DDI-W and Contra-W weight each safety relation by its observed frequency.
Lower values indicate safer predictions for all the generated medication sets.  Full metric definitions are
provided in Appendix~\ref{app:evaluation_metrics}.
\section{Results}

\begin{table*}[t]
    \centering
    \caption{
    Prediction performance across MIMIC-IV and MIMIC-III. Within each LLM backbone group, \textbf{bold} marks the best and \underline{underline} marks the second-best (excluding the Avg. \#Pred column).
    Avg. \#Pred reports the average number of predicted medications per case and is compared with the average ground-truth size.
    }
    \label{tab:prediction}
    \scriptsize
    \setlength{\tabcolsep}{3.5pt}
    \begin{tabular}{lccccc|ccccc}
    \toprule
    \multirow{2}{*}{\textbf{Method}}
    & \multicolumn{5}{c|}{\textbf{MIMIC-IV}}
    & \multicolumn{5}{c}{\textbf{MIMIC-III}} \\
    \cmidrule(lr){2-6} \cmidrule(lr){7-11}
    & \textbf{Precision} & \textbf{Recall} & \textbf{F1} & \textbf{Jaccard} & \textbf{Avg. \#Pred}
    & \textbf{Precision} & \textbf{Recall} & \textbf{F1} & \textbf{Jaccard} & \textbf{Avg. \#Pred} \\
    \midrule

    \multicolumn{11}{c}{\cellcolor{gray!15}\textit{Deep learning medication recommendation baselines}} \\
    GAMENet & 0.2510 & 0.5062 & 0.3356 & 0.2171 & 23.92 & 0.1521 & 0.3684 & 0.2153 & 0.1834 & 18.16 \\
    SafeDrug & 0.2639 & 0.5280 & 0.3519 & 0.2286 & 16.53 & 0.1638 & 0.4227 & 0.2361 & 0.1914 & 13.29\\
    DrugDoctor
    & 0.2796 & 0.5190 & 0.3634 & 0.2369 & 17.17
    & 0.1551 & 0.3915 & 0.2232 & 0.1909 & 13.88 \\
    KEHGCN & 0.2980 & 0.5493 & 0.3864 & 0.2548 & 21.31 & 0.1634 & 0.4284 & 0.2365 & 0.2018 & 16.75 \\
    \midrule

    \multicolumn{11}{c}{\cellcolor{gray!15}\textit{Open-source Medical LLM baselines (direct prompting)}} \\
    UltraMedical-70B
    & 0.2992 & 0.2402 & 0.2665 & 0.1493 & 7.36
    & 0.1254 & 0.0999 & 0.1112 & 0.0486 & 4.27 \\
    Med42-v2-70B
    & 0.2414 & 0.2220 & 0.2313 & 0.1273 & 8.44
    & 0.1293 & 0.1245 & 0.1269 & 0.0689 & 5.16 \\
    Llama3.1-Aloe-Beta-70B
    & 0.3234 & 0.2370 & 0.2735 & 0.1572 & 6.72
    & 0.1499 & 0.1098 & 0.1268 & 0.0708 & 3.92 \\
    HuatuoGPT-o1
    & 0.3100 & 0.1417 & 0.1945 & 0.1085 & 4.19
    & 0.1476 & 0.1162 & 0.1300 & 0.0715 & 4.22 \\
    OpenBioLLM-70B
    & 0.4104 & 0.3691 & 0.3887 & 0.2481 & 8.25
    & 0.1699 & 0.0754 & 0.1045 & 0.0529 & 2.38 \\
    \midrule

    \multicolumn{11}{c}{\cellcolor{gray!15}\textit{Closed-source LLM: GPT-5.2}} \\
    Direct Prompting
    & 0.3528 & \textbf{0.4447} & 0.3935 & 0.2535 & 11.62
    & 0.1891 & \textbf{0.3383} & 0.2426 & 0.1424 & 9.57 \\
    General Agent
    & 0.4271 & 0.3562 & 0.3884 & 0.2475 & 8.50
    & 0.2087 & \underline{0.3215} & 0.2531 & 0.1526 & 8.24 \\
    RareAgents
    & 0.3902 & \underline{0.4072} & \underline{0.3985} & \underline{0.2573} & 9.61
    & 0.1586 & 0.2019 & 0.1776 & 0.1031 & 6.81 \\
    Ours (w/o safety filter)
    & \underline{0.4489} & 0.3837 & \textbf{0.4138} & 0.2570 & 8.71
    & \underline{0.2991} & 0.2654 & \textbf{0.2812} & \textbf{0.1702} & 4.70 \\
    \rowcolor{blue!8}
    Ours (w/ safety filter)
    & \textbf{0.5210} & 0.3160 & 0.3934 & \textbf{0.2631} & 6.18
    & \textbf{0.3145} & 0.2277 & \underline{0.2641} & \underline{0.1620} & 3.85 \\
    \midrule

    \multicolumn{11}{c}{\cellcolor{gray!15}\textit{Closed-source LLM: Claude Sonnet-4.6}} \\
    Direct Prompting
    & 0.2650 & \textbf{0.5662} & 0.3610 & 0.2100 & 19.70
    & 0.1328 & 0.4305 & 0.2029 & 0.1103 & 17.25 \\
    General Agent
    & 0.3016 & 0.5152 & 0.3804 & 0.2229 & 17.41
    & 0.1513 & \underline{0.4323} & 0.2240 & 0.1250 & 15.20 \\
    RareAgents
    & \textbf{0.3656} & 0.4343 & 0.3970 & 0.2531 & 10.94
    & 0.1260 & 0.2152 & 0.1590 & 0.0873 & 9.56 \\
    Ours (w/o safety filter)
    & 0.3557 & \underline{0.5515} & \textbf{0.4325} & \textbf{0.2650} & 15.80
    & \underline{0.1950} & \textbf{0.4711} & \textbf{0.2758} & \textbf{0.1586} & 12.37 \\
    \rowcolor{blue!8}
    Ours (w/ safety filter)
    & \underline{0.3611} & 0.5103 & \underline{0.4229} & \underline{0.2617} & 14.40
    & \textbf{0.1971} & 0.4125 & \underline{0.2667} & \underline{0.1528} & 10.72 \\
    \midrule

    \multicolumn{11}{c}{\cellcolor{gray!15}\textit{Open-source LLM: Qwen3-32B}} \\
    Direct Prompting
    & 0.3452 & 0.3145 & 0.3292 & 0.2017 & 8.36
    & 0.1381 & 0.1086 & 0.1216 & 0.0616 & 4.22 \\
    General Agent
    & 0.3717 & 0.2978 & 0.3245 & 0.2030 & 6.48
    & 0.1828 & 0.1067 & 0.1350 & 0.0700 & 3.14 \\
    RareAgents
    & \underline{0.3724} & 0.2931 & 0.3280 & 0.2011 & 7.24
    & 0.1612 & 0.0806 & 0.1074 & 0.0552 & 2.74 \\
    Ours (w/o safety filter)
    & \textbf{0.3939} & \underline{0.3277} & \textbf{0.3578} & \textbf{0.2276} & 7.38
    & \textbf{0.2899} & \textbf{0.2712} & \textbf{0.2514} & \textbf{0.1569} & 4.53 \\
    \rowcolor{blue!8}
    Ours (w/ safety filter)
    & 0.3576 & \textbf{0.3418} & \underline{0.3495} & \underline{0.2212} & 6.51
    & \underline{0.2384} & \underline{0.2533} & \underline{0.2266} & \underline{0.1265} & 3.44 \\
    \midrule

    \multicolumn{11}{c}{\cellcolor{gray!15}\textit{Open-source LLM: Gemma3-27B-IT}} \\
    Direct Prompting
    & 0.2795 & \underline{0.3673} & 0.3174 & 0.1856 & 12.06
    & 0.0986 & 0.1477 & 0.1183 & 0.0634 & 8.03 \\
    General Agent
    & 0.3073 & 0.3595 & 0.3313 & 0.1984 & 10.7
    & 0.1349 & 0.1411 & 0.1379 & 0.0723 & 5.63 \\
    RareAgents
    & 0.3677 & 0.3050 & 0.3334 & 0.2006 & 7.61
    & 0.1041 & 0.0773 & 0.0887 & 0.0462 & 3.98 \\
    Ours (w/o safety filter)
    & \underline{0.3789} & \textbf{0.3978} & \textbf{0.3882} & \underline{0.2446} & 9.65
    & \underline{0.2289} & \textbf{0.3121} & \textbf{0.2641} & \textbf{0.1537} & 7.30 \\
    \rowcolor{blue!8}
    Ours (w/ safety filter)
    & \textbf{0.4374} & 0.3239 & \underline{0.3722} & \textbf{0.2512} & 6.79
    & \textbf{0.2479} & \underline{0.2567} & \underline{0.2146} & \underline{0.1146} & 5.39 \\
    \bottomrule
    \end{tabular}
\end{table*}

\subsection{Quantitative Results}
Table~\ref{tab:prediction} reports medication recommendation performance on
MIMIC-IV and MIMIC-III. \ours{} consistently outperforms direct prompting and
generic agent baselines across all backbones, achieving the best pre-filter F1
on MIMIC-IV for every model tested. The safety filter trades a small drop in
recall for higher precision and smaller predicted sets, as expected from a
verifier that removes unsupported candidates. Without task-specific training,
\ours{} also matches or exceeds supervised baselines on F1 and Jaccard.
MIMIC-III shows the same pattern, confirming that the workflow generalizes
beyond MIMIC-IV.

\begin{table*}[t]
    \centering
    \caption{
    Safety analysis of predicted medication sets across MIMIC-IV and MIMIC-III. All safety metrics are lower-is-better. Within each LLM backbone group, \textbf{bold} marks the best and \underline{underline} marks the second-best in each column (excluding the Ground Truth row and the Avg. \#Pred column).
    DDI-B and Contra-B report the fraction of cases with at least one DDI or
contraindication, while DDI-W and Contra-W report their frequency-weighted
rates. Avg. \#Pred is the average number of predicted ATC codes per case.
    }
    \label{tab:safety}
    \scriptsize
    \setlength{\tabcolsep}{3.5pt}
    \begin{tabular}{lccccc|ccccc}
    \toprule
    \multirow{2}{*}{\textbf{Method}}
    & \multicolumn{5}{c|}{\textbf{MIMIC-IV} ($\downarrow$ except \#Pred)}
    & \multicolumn{5}{c}{\textbf{MIMIC-III} ($\downarrow$ except \#Pred)} \\
    \cmidrule(lr){2-6} \cmidrule(lr){7-11}
    & \textbf{DDI-B}
    & \textbf{DDI-W}
    & \textbf{Contra-B}
    & \textbf{Contra-W}
    & \textbf{Avg. \#Pred}
    & \textbf{DDI-B}
    & \textbf{DDI-W}
    & \textbf{Contra-B}
    & \textbf{Contra-W}
    & \textbf{Avg. \#Pred} \\
    \midrule
    
    \rowcolor{gray!15}
    \multicolumn{11}{c}{\textit{Ground-truth reference}} \\
    Ground Truth
    & 0.3013 & 0.1843 & 0.0043 & 0.0011 & 9.17
    & 0.2460 & 0.1507 & 0.0023 & 0.0007 & 5.36 \\
    \midrule
    
    \rowcolor{gray!15}
    \multicolumn{11}{c}{\textit{Deep learning medication recommendation baselines}} \\
    GAMENet & 0.7516 & 0.6023 & 0.0165 & 0.0074 & 23.92
    & 0.6238 & 0.5848 & 0.0081 & 0.0030 & 18.16 \\
    SafeDrug & 0.5869 & 0.4326 & 0.0083 & 0.0035 & 16.53
    & 0.4055 & 0.3268 & 0.0053 & 0.0015 & 13.29\\
    % DrugDoctor (threshold=0.2)
    % & 0.9413 & 0.8771 & 0.0255 & 0.0078 & 30.65
    % & 0.7444 & 0.6835 & 0.0117 & 0.0032 & 21.25 \\
    DrugDoctor
    & 0.7010 & 0.5867 & 0.0157 & 0.0049 & 17.17
    & 0.5727 & 0.5058 & 0.0075 & 0.0022 & 13.88 \\
    KEHGCN & 0.6134 & 0.4653 & 0.0072 & 0.0031 & 21.31 & 0.4351 & 0.3044 & 0.0042 & 0.0013 & 16.75 \\
    % UDC Health (threshold=0.2)
    % & 0.4948 & 0.3795 & 0.0081 & 0.0014 & 11.63
    % & --     & --     & --     & --     & -- \\
    % UDC Health (threshold=0.4)
    % & 0.1518 & 0.1055 & 0.0038 & 0.0007 & 5.62
    % & --     & --     & --     & --     & -- \\
    \midrule

    \multicolumn{11}{c}{\cellcolor{gray!15}\textit{Medical LLM baselines (direct prompting)}} \\
    UltraMedical-70B
    & 0.4033 & 0.3132 & 0.0058 & 0.0016 & 7.36
    & 0.3356 & 0.2644 & 0.0043 & 0.0015 & 4.27 \\
    Med42-v2-70B
    & 0.5477 & 0.4199 & 0.0075 & 0.0024 & 8.44
    & 0.5206 & 0.3922 & 0.0063 & 0.0021 & 5.16 \\
    Llama3.1-Aloe-Beta-70B
    & 0.4245 & 0.3092 & 0.0056 & 0.0015 & 6.72
    & 0.3840 & 0.2685 & 0.0056 & 0.0016 & 3.92 \\
    HuatuoGPT-o1
    & 0.2330 & 0.1592 & 0.0035 & 0.0011 & 4.19
    & 0.3503 & 0.2376 & 0.0046 & 0.0015 & 4.22 \\
    OpenBioLLM-70B
    & 0.3878 & 0.2948 & 0.0054 & 0.0013 & 8.25
    & 0.2453 & 0.1732 & 0.0028 & 0.0009 & 2.38 \\
    \midrule

    \rowcolor{gray!15}
    \multicolumn{11}{c}{\textit{Closed-source LLM: GPT-5.2}} \\
    Direct Prompting
    & 0.7218 & 0.5914 & 0.0067 & 0.0014 & 11.62
    & 0.9022 & 0.8360 & 0.0010 & \underline{0.0001} & 9.57 \\
    General Agent
    & \underline{0.4896} & 0.3902 & 0.0060 & 0.0013 & 8.50
    & 0.8686 & 0.7773 & 0.0010 & 0.0004 & 8.24 \\
    RareAgents
    & 0.4980 & \underline{0.3650} & \underline{0.0030} & \textbf{0.0006} & 9.61
    & 0.6101 & 0.5214 & \underline{0.0004} & \underline{0.0001} & 6.81 \\
    Ours (w/o safety filter)
    & 0.5340 & 0.4126 & 0.0078 & 0.0017 & 8.71
    & \underline{0.5765} & \underline{0.4833} & 0.0050 & 0.0007 & 4.70 \\
    \rowcolor{blue!8}
    Ours (w/ safety filter)
    & \textbf{0.2819} & \textbf{0.2115} & \textbf{0.0028} & \underline{0.0007} & 6.18
    & \textbf{0.4386} & \textbf{0.3572} & \textbf{0.0000} & \textbf{0.0000} & 3.85 \\
    \midrule

    \rowcolor{gray!15}
    \multicolumn{11}{c}{\textit{Closed-source LLM: Claude Sonnet-4.6}} \\
    Direct Prompting
    & 0.9408 & 0.8668 & 0.0106 & 0.0026 & 19.70
    & 0.9875 & 0.9643 & 0.0140 & 0.0031 & 17.25 \\
    General Agent
    & 0.8951 & 0.8061 & 0.0106 & 0.0032 & 17.41
    & 0.9886 & 0.9582 & 0.0134 & 0.0029 & 15.20 \\
    RareAgents
    & \textbf{0.6162} & \textbf{0.4952} & \underline{0.0043} & \underline{0.0010} & 10.94
    & \textbf{0.8007} & \textbf{0.6990} & \textbf{0.0000} & \textbf{0.0000} & 9.56 \\
    Ours (w/o safety filter)
    & 0.9096 & 0.8283 & 0.0076 & 0.0013 & 15.80
    & 0.9679 & 0.9403 & 0.0048 & 0.0012 & 12.37 \\
    \rowcolor{blue!8}
    Ours (w/ safety filter)
    & \underline{0.7981} & \underline{0.7068} & \textbf{0.0030} & \textbf{0.0009} & 14.40
    & \underline{0.9376} & \underline{0.8877} & \underline{0.0006} & \underline{0.0003} & 10.72 \\
    \midrule

    \rowcolor{gray!15}
    \multicolumn{11}{c}{\textit{Open-source LLM: Qwen3-32B}} \\
    Direct Prompting
    & 0.5318 & 0.4095 & 0.0074 & 0.0018 & 8.36
    & 0.4243 & 0.3168 & 0.0038 & 0.0011 & 4.22 \\
    General Agent
    & 0.4100 & \underline{0.3185} & 0.0050 & 0.0013 & 6.48
    & \underline{0.3249} & \underline{0.2389} & \underline{0.0030} & \underline{0.0007} & 3.14 \\
    RareAgents
    & \underline{0.3959} & \textbf{0.3107} & 0.0077 & 0.0022 & 7.24
    & \textbf{0.1730} & \textbf{0.1154} & 0.0046 & 0.0013 & 2.74 \\
    Ours (w/o safety filter)
    & 0.5262 & 0.4177 & \underline{0.0044} & \underline{0.0011} & 7.38
    & 0.7396 & 0.4698 & 0.0090 & 0.0023 & 4.53 \\
    \rowcolor{blue!8}
    Ours (w/ safety filter)
    & \textbf{0.3585} & 0.3758 & \textbf{0.0037} & \textbf{0.0007} & 6.51
    & 0.3927 & 0.3513 & \textbf{0.0000} & \textbf{0.0000} & 3.44 \\
    \midrule

    \rowcolor{gray!15}
    \multicolumn{11}{c}{\textit{Open-source LLM: Gemma3-27B-IT}} \\
    Direct Prompting
    & 0.6960 & 0.5747 & 0.0095 & 0.0022 & 12.06
    & 0.7643 & 0.6441 & 0.0061 & 0.0015 & 8.03 \\
    General Agent
    & 0.6620 & 0.5530 & 0.0088 & 0.0022 & 10.7
    & 0.6296 & 0.5304 & 0.0065 & 0.0019 & 5.63 \\
    RareAgents
    & \underline{0.4753} & \underline{0.3624} & \underline{0.0069} & \underline{0.0017} & 7.61
    & \underline{0.4876} & \underline{0.3828} & \underline{0.0028} & \underline{0.0010} & 3.98 \\
    Ours (w/o safety filter)
    & 0.6683 & 0.5761 & 0.0086 & \underline{0.0017} & 9.65
    & 0.8170 & 0.7268 & 0.0062 & 0.0013 & 7.30 \\
    \rowcolor{blue!8}
    Ours (w/ safety filter)
    & \textbf{0.3695} & \textbf{0.2865} & \textbf{0.0030} & \textbf{0.0007} & 6.79
    & \textbf{0.3650} & \textbf{0.3258} & \textbf{0.0000} & \textbf{0.0000} & 5.39 \\
    \bottomrule
    \end{tabular}
\end{table*}

\subsection{Drug Safety Analysis}
\label{sec:drug_safety}
We evaluate medication recommendation safety by comparing each predicted ATC-L4 set with the DDI and contraindication matrices constructed from external safety resources (Table~\ref{tab:safety}). Deep learning baselines and unfiltered LLM agents often generate larger medication sets with higher DDI and contraindication rates. Therefore, these methods cannot reliably control medication safety risks without explicit safety verification.
Our safety filter consistently reduces DDI and contraindication rates while keeping the predicted set size closer to the ground-truth set size. The largest reductions are observed for GPT-5.2 and Gemma3-27B-IT, where filtered predictions show safety profiles that are better aligned with the extracted safety knowledge. Overall, these findings suggest that reliable fine-grained medication recommendation requires a specific safety verification instead of relying on supervised learning loss or LLM generation alone. 

\subsection{Ablation Studies}
\label{sec:ablation}

To quantify the marginal contribution of each component in the generation pipeline, we ablate \ours{} by removing one component at a time while keeping all other settings fixed. All ablation experiments are conducted on MIMIC-IV using Gemma3-27B-IT (Table~\ref{tab:ablation}); subsequent analyses use the same setting. Replacing the specialty panel with a general agent causes the largest F1 drop
($-$0.057) and increases the predicted set size (10.7 vs.\ 9.65), confirming that specialty routing constrains each expert to its scope. Removing the critique hurts precision most (0.379$\to$0.322) while recall stays flat, consistent with its role as a pruning stage. Removing the medical resource instead hurts recall most (0.398$\to$0.325), indicating that retrieved evidence helps recover medications not surfaced by the LLM alone.

\begin{table*}[t]
    \footnotesize
    \centering
    \setlength{\tabcolsep}{5pt}
    \caption{\textbf{Prediction ablations.} Each variant removes one component while keeping other settings fixed. \textcolor{red!70!black}{Red} marks decreases. \#Pred deltas are shown in black since smaller or larger sets are not directly better or worse.}
    \label{tab:ablation}
    \begin{tabular}{lccccc}
    \toprule
    \textbf{Method}
    & \textbf{Precision}
    & \textbf{Recall}
    & \textbf{F1}
    & \textbf{Jaccard}
    & \textbf{\#Pred} \\
    \midrule
    Full \ours & 0.3789 & 0.3978 & 0.3882 & 0.2446 & 9.65 \\
    w/o Specialty Experts
    & 0.3073\,\textcolor{red!70!black}{($-$0.0716)}
    & 0.3595\,\textcolor{red!70!black}{($-$0.0383)}
    & 0.3313\,\textcolor{red!70!black}{($-$0.0569)}
    & 0.1984\,\textcolor{red!70!black}{($-$0.0462)}
    & 10.7\,($+$1.05) \\
    w/o Summarizer
    & 0.3564\,\textcolor{red!70!black}{($-$0.0225)}
    & 0.3795\,\textcolor{red!70!black}{($-$0.0183)}
    & 0.3679\,\textcolor{red!70!black}{($-$0.0203)}
    & 0.2272\,\textcolor{red!70!black}{($-$0.0174)}
    & 9.28\,($-$0.37) \\
    w/o Medical Resource
    & 0.3699\,\textcolor{red!70!black}{($-$0.0090)}
    & 0.3249\,\textcolor{red!70!black}{($-$0.0729)}
    & 0.3673\,\textcolor{red!70!black}{($-$0.0209)}
    & 0.2231\,\textcolor{red!70!black}{($-$0.0215)}
    & 9.57\,($-$0.08) \\
    w/o Critique
    & 0.3225\,\textcolor{red!70!black}{($-$0.0564)}
    & 0.3849\,\textcolor{red!70!black}{($-$0.0129)}
    & 0.3589\,\textcolor{red!70!black}{($-$0.0293)}
    & 0.2370\,\textcolor{red!70!black}{($-$0.0076)}
    & 9.71\,($+$0.06) \\
    \bottomrule
    \end{tabular}
\end{table*}

\subsection{Diagnostic Analysis}
\label{sec:diagnostic_analysis}

\begin{figure*}[t]
    \centering
    \begin{subfigure}[t]{0.32\textwidth}
        \centering
        \includegraphics[width=\linewidth]{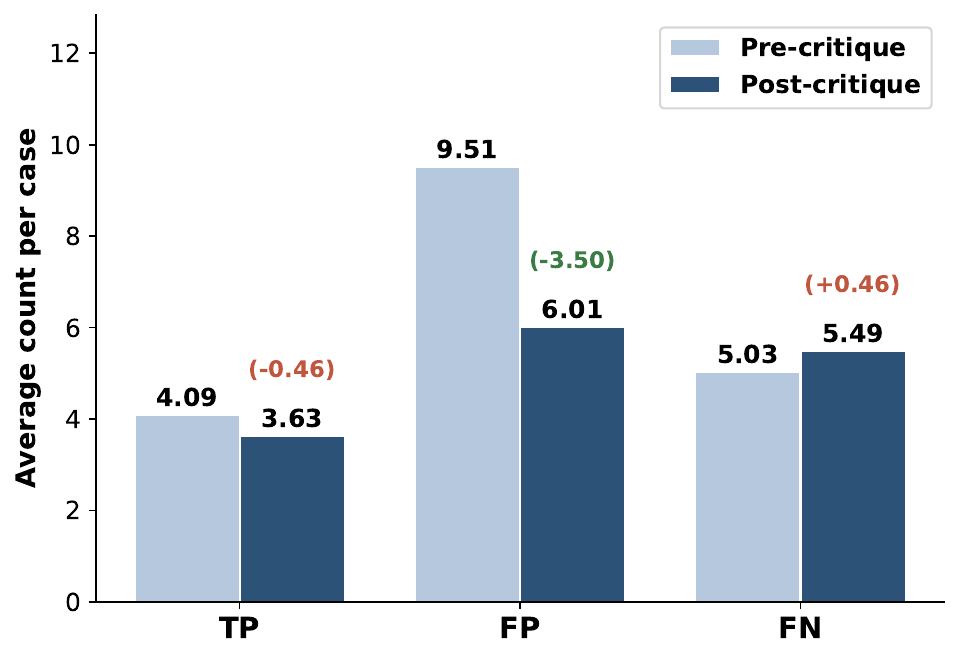}
        \caption{Effect of critique on TP/FP/FN.}
        \label{fig:critique_tpfpfn}
    \end{subfigure}
    \hfill
    \begin{subfigure}[t]{0.32\textwidth}
        \centering
        \includegraphics[width=\linewidth]{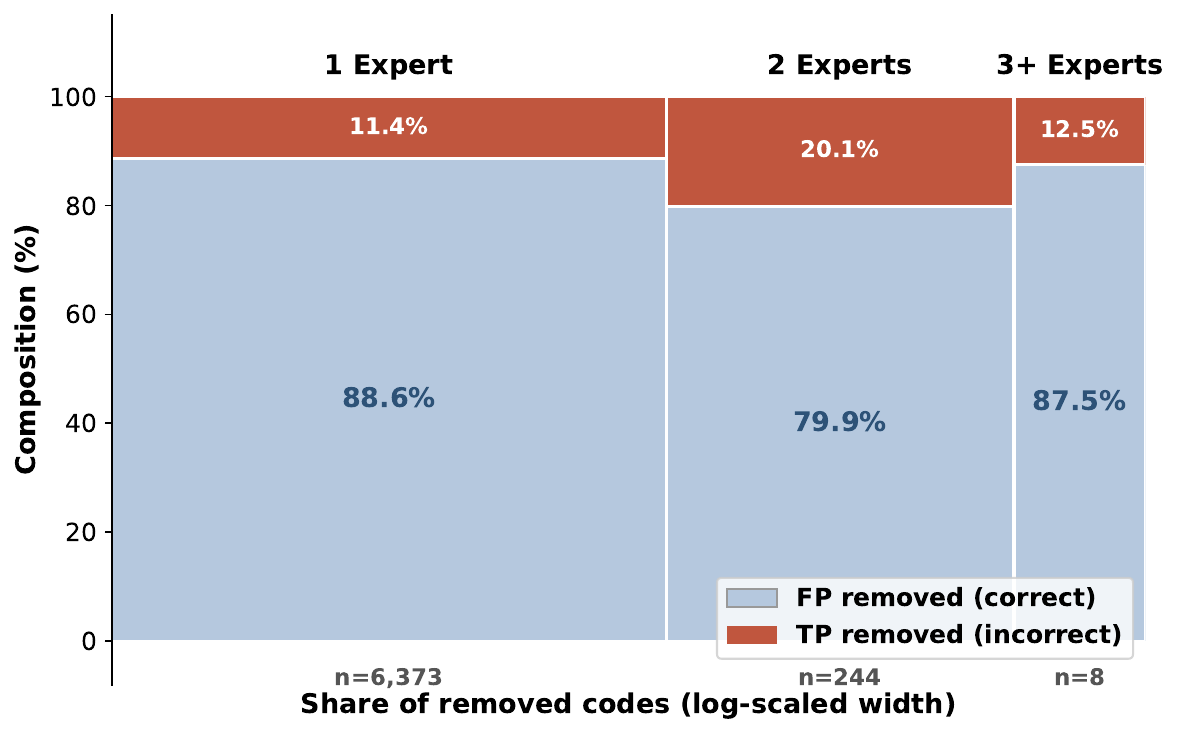}
        \caption{Removed meds by expert support.}
        \label{fig:critique_support}
    \end{subfigure}
    \hfill
    \begin{subfigure}[t]{0.32\textwidth}
        \centering
        \includegraphics[width=\linewidth]{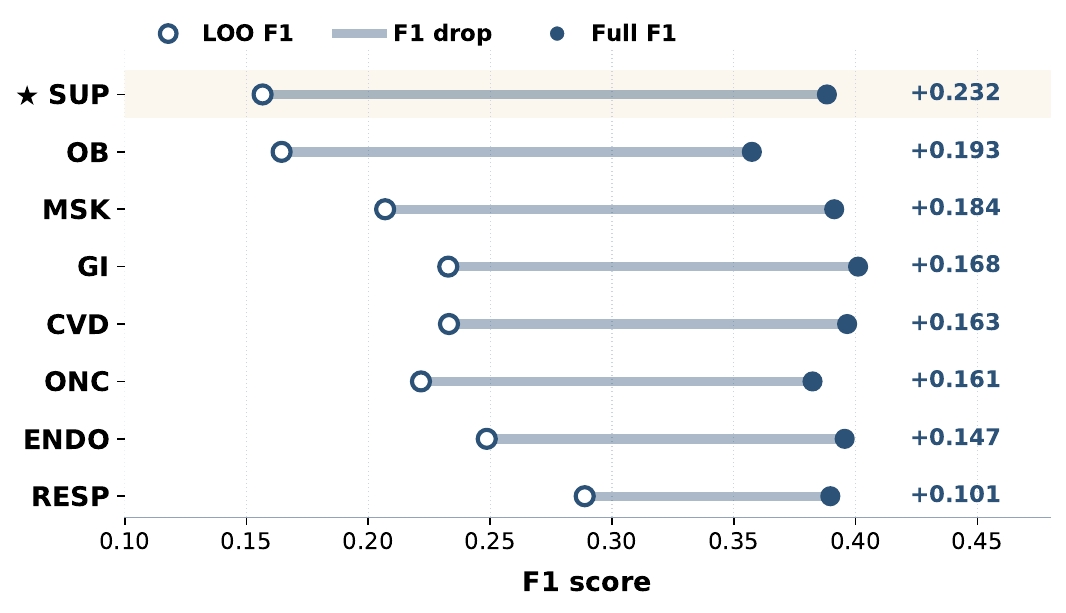}
        \caption{LOO subgroup performance.}
        \label{fig:expert_subgroup}
    \end{subfigure}
    \caption{\textbf{Diagnostic and expert analyses.}
    (a)~Critique substantially reduces average false positives.
    (b)~Critique primarily removes medications proposed by only one expert, and effectively removes false-positives.
    (c)~Leave-one-expert-out subgroup performance; expert abbreviations are defined at the beginning of Appendix~\ref{app:expert_analysis}.}
    \label{fig:diagnostic_expert}
\end{figure*}

We analyze how the critique stage reshapes the proposed medication set by comparing pre- and post-critique predictions.
Figure~\ref{fig:critique_tpfpfn} shows that critique reduces false positives
from 9.51 to 6.01 per case at a cost of only 0.46 true positives, acting as an effective precision-oriented pruning stage.
Figure~\ref{fig:critique_support} further reveals why: of the 6,625 removed medications, 6,373 (96.2\%) were proposed by only a single expert, while medications corroborated by multiple specialists were rarely pruned. This suggests that the critique operates conservatively, selectively targeting candidates that lack independent corroboration rather than indiscriminately reducing the set size.
Supplementary diagnostic results are reported in Appendix~\ref{app:diagnostic_analysis}; and qualitative case studies are presented in Appendix~\ref{app:case_analysis}.

\subsection{Expert Analysis}
\label{sec:expert_analysis}
To test whether individual experts provide marginal value beyond the panel as a whole, we conduct a
leave-one-expert-out (LOO) analysis on activation-defined subgroups. For each expert, we compare subgroup F1 before and after removing only
that expert. Figure~\ref{fig:expert_subgroup} shows that every activated
expert contributes positively, with F1 drops ranging from +0.101 (RESP) to +0.232 (SUP). The universal supportive-care expert contributes the most, consistent with its broad activation across cases. Even the least impactful expert (RESP) still yields a meaningful F1 gain, indicating that no activated expert is redundant.
Appendix~\ref{app:expert_analysis} reports activation sparsity and
per-expert TP contributions.

\subsection{Medication Granularity and Safety Evaluation}
\label{sec:granularity_safety}

\begin{figure}[t]
    \centering
    \includegraphics[width=\columnwidth]{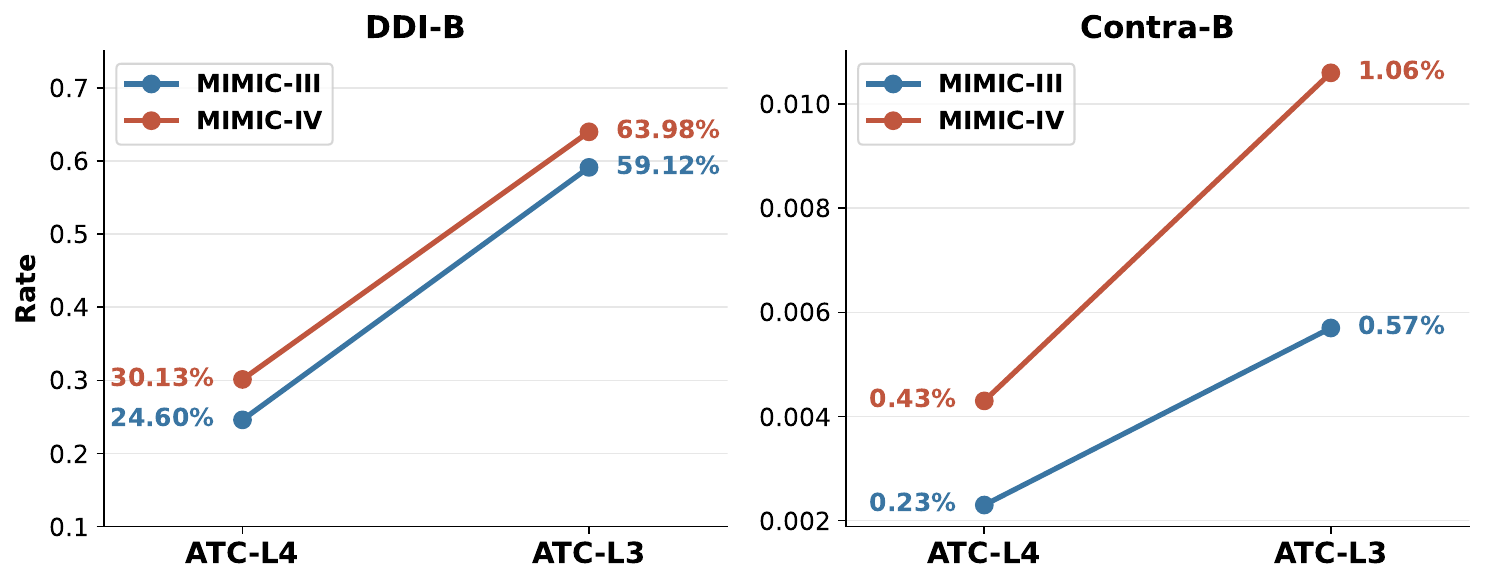}
    \caption{Binary DDI and contraindication rates under ATC-L4 and ATC-L3 medication vocabulary.}
    \label{fig:atc_level_safety_slope}
    \vspace{-0.5em}
\end{figure}

To quantify how medication granularity affects safety evaluation, we compare binary DDI and contraindication rates under ATC-L4 and ATC-L3 representations. We compute the safety metrics directly from ATC-L4 medication sets. We then collapse each ATC-L4 code to its ATC-L3 parent and recompute these metrics at the ATC-L3 level. 
Figure~\ref{fig:atc_level_safety_slope} shows that ATC-L3 yields higher DDI and contraindication rates than ATC-L4. On MIMIC-III, DDI-B increases from 24.60\% to 59.12\%, and Contra-B increases from 0.23\% to 0.57\%. On MIMIC-IV, DDI-B increases from 30.13\% to 63.98\%, and Contra-B increases from 0.43\% to 1.06\%. This finding supports fine-grained medication recommendation at ATC-L4, which preserves medication subgroup distinctions and avoids overestimation of safety risk.

\subsection{Efficiency Analysis}
\label{sec:efficiency}

We evaluate the inference efficiency of \ours on MIMIC-IV in terms of  average LLM calls, input/output tokens, and per-case. As shown in Table~\ref{tab:efficiency},   despite its evidence-grounded multi-agent workflow, \ours{} uses fewer LLM calls than RareAgents (9.4 vs.\ 10.1), suggesting that sparse routing avoids
unnecessary expert calls. The final safety filter is lightweight: relative to the no-filter variant, it
increases latency from 27.7 to 29.5 seconds per case and adds only one LLM call. A per-phase breakdown is provided in Appendix~\ref{app:efficiency_breakdown}.

\begin{table}[t]
\centering
\scriptsize
\setlength{\tabcolsep}{5pt}
\caption{\textbf{Efficiency analysis.}
We report average LLM calls, token usage, and wall-clock latency per case.}
\label{tab:efficiency}
\begin{tabular}{@{}lrrrr@{}}
\toprule
\textbf{Method} & \textbf{\#Calls} & \textbf{In Tok.} & \textbf{Out Tok.} & \textbf{Sec./case} \\
\midrule
Direct Prompting & 1.0 & 885 & 517 & 9.6 \\
General Agent & 3.0 & 4,928 & 788 & 10.2 \\
RareAgents & 10.1 & 13,478 & 1,710 & 13.3 \\
\ours w/o safety filter & 8.4 & 24,260 & 5,988 & 27.7 \\
\ours w/ safety filter & 9.4 & 24,847 & 6,070 & 29.5 \\
\bottomrule
\end{tabular}
\end{table}
\section{Conclusion}

We presented \ours{}, a knowledge-grounded multi-agent framework for safe and explainable medication recommendation. We introduce a fine-grained ATC-L4 setting that preserves clinically important subgroup distinctions often hidden by coarser benchmarks. \ours{} routes each case to multiple specialty experts, grounds generation in patient context and indication evidence, and applies safety verification using DDI and contraindication resources. Experiments on MIMIC-III and MIMIC-IV show that \ours{} improves prediction accuracy over deep learning, LLM, and agentic baselines while reducing safety risks and controlling predicted set size. Ablations further show that routing, summarization, evidence grounding, critique, and safety verification provide complementary gains. We hope this work encourages more verifiable agentic systems for high-stakes clinical NLP.

\clearpage

\section*{Limitations}

SafeRx-Agent evaluates medication safety using mapped external resources for drug interactions and disease-conditioned contraindications. These resource-grounded metrics systematically assess safety risks in generated medication sets, while their coverage depends on the underlying knowledge sources and ATC-L4 mapping. Our safety and explainability analyses rely on structured evidence checks and traceable rationales, and future work may incorporate clinician review to assess their clinical validity. Our evaluation is retrospective and offline; deploying the system in real clinical workflows would require prospective validation and physician oversight to ensure patient safety.

\bibliography{custom}

@misc{liu2025largelanguagemodeldistilling,
      title={Large Language Model Distilling Medication Recommendation Model}, 
      author={Qidong Liu and Xian Wu and Xiangyu Zhao and Yuanshao Zhu and Zijian Zhang and Feng Tian and Yefeng Zheng},
      year={2025},
      eprint={2402.02803},
      archivePrefix={arXiv},
      primaryClass={cs.IR},
      url={https://arxiv.org/abs/2402.02803}, 
}

@article{drugdoctor,
    author = {Kuang, Yabin and Xie, Minzhu},
    title = {DrugDoctor: enhancing drug recommendation in cold-start scenario via visit-level representation learning and training},
    journal = {Briefings in Bioinformatics},
    volume = {25},
    number = {6},
    pages = {bbae464},
    year = {2024},
    month = {09},
    issn = {1477-4054},
    doi = {10.1093/bib/bbae464},
    url = {https://doi.org/10.1093/bib/bbae464},
    eprint = {https://academic.oup.com/bib/article-pdf/25/6/bbae464/59242513/bbae464.pdf},
}

@misc{med42v2,
      title={Med42-v2: A Suite of Clinical LLMs}, 
      author={Clément Christophe and Praveen K Kanithi and Tathagata Raha and Shadab Khan and Marco AF Pimentel},
      year={2024},
      eprint={2408.06142},
      archivePrefix={arXiv},
      primaryClass={cs.CL},
      url={https://arxiv.org/abs/2408.06142}, 
}

@misc{gemma3,
      title={Gemma 3 Technical Report}, 
      author={Gemma Team and Aishwarya Kamath and Johan Ferret and Shreya Pathak and Nino Vieillard and Ramona Merhej and Sarah Perrin and Tatiana Matejovicova and Alexandre Ramé and Morgane Rivière and Louis Rouillard and Thomas Mesnard and Geoffrey Cideron and Jean-bastien Grill and Sabela Ramos and Edouard Yvinec and Michelle Casbon and Etienne Pot and Ivo Penchev and Gaël Liu and Francesco Visin and Kathleen Kenealy and Lucas Beyer and Xiaohai Zhai and Anton Tsitsulin and Robert Busa-Fekete and Alex Feng and Noveen Sachdeva and Benjamin Coleman and Yi Gao and Basil Mustafa and Iain Barr and Emilio Parisotto and David Tian and Matan Eyal and Colin Cherry and Jan-Thorsten Peter and Danila Sinopalnikov and Surya Bhupatiraju and Rishabh Agarwal and Mehran Kazemi and Dan Malkin and Ravin Kumar and David Vilar and Idan Brusilovsky and Jiaming Luo and Andreas Steiner and Abe Friesen and Abhanshu Sharma and Abheesht Sharma and Adi Mayrav Gilady and Adrian Goedeckemeyer and Alaa Saade and Alex Feng and Alexander Kolesnikov and Alexei Bendebury and Alvin Abdagic and Amit Vadi and András György and André Susano Pinto and Anil Das and Ankur Bapna and Antoine Miech and Antoine Yang and Antonia Paterson and Ashish Shenoy and Ayan Chakrabarti and Bilal Piot and Bo Wu and Bobak Shahriari and Bryce Petrini and Charlie Chen and Charline Le Lan and Christopher A. Choquette-Choo and CJ Carey and Cormac Brick and Daniel Deutsch and Danielle Eisenbud and Dee Cattle and Derek Cheng and Dimitris Paparas and Divyashree Shivakumar Sreepathihalli and Doug Reid and Dustin Tran and Dustin Zelle and Eric Noland and Erwin Huizenga and Eugene Kharitonov and Frederick Liu and Gagik Amirkhanyan and Glenn Cameron and Hadi Hashemi and Hanna Klimczak-Plucińska and Harman Singh and Harsh Mehta and Harshal Tushar Lehri and Hussein Hazimeh and Ian Ballantyne and Idan Szpektor and Ivan Nardini and Jean Pouget-Abadie and Jetha Chan and Joe Stanton and John Wieting and Jonathan Lai and Jordi Orbay and Joseph Fernandez and Josh Newlan and Ju-yeong Ji and Jyotinder Singh and Kat Black and Kathy Yu and Kevin Hui and Kiran Vodrahalli and Klaus Greff and Linhai Qiu and Marcella Valentine and Marina Coelho and Marvin Ritter and Matt Hoffman and Matthew Watson and Mayank Chaturvedi and Michael Moynihan and Min Ma and Nabila Babar and Natasha Noy and Nathan Byrd and Nick Roy and Nikola Momchev and Nilay Chauhan and Noveen Sachdeva and Oskar Bunyan and Pankil Botarda and Paul Caron and Paul Kishan Rubenstein and Phil Culliton and Philipp Schmid and Pier Giuseppe Sessa and Pingmei Xu and Piotr Stanczyk and Pouya Tafti and Rakesh Shivanna and Renjie Wu and Renke Pan and Reza Rokni and Rob Willoughby and Rohith Vallu and Ryan Mullins and Sammy Jerome and Sara Smoot and Sertan Girgin and Shariq Iqbal and Shashir Reddy and Shruti Sheth and Siim Põder and Sijal Bhatnagar and Sindhu Raghuram Panyam and Sivan Eiger and Susan Zhang and Tianqi Liu and Trevor Yacovone and Tyler Liechty and Uday Kalra and Utku Evci and Vedant Misra and Vincent Roseberry and Vlad Feinberg and Vlad Kolesnikov and Woohyun Han and Woosuk Kwon and Xi Chen and Yinlam Chow and Yuvein Zhu and Zichuan Wei and Zoltan Egyed and Victor Cotruta and Minh Giang and Phoebe Kirk and Anand Rao and Kat Black and Nabila Babar and Jessica Lo and Erica Moreira and Luiz Gustavo Martins and Omar Sanseviero and Lucas Gonzalez and Zach Gleicher and Tris Warkentin and Vahab Mirrokni and Evan Senter and Eli Collins and Joelle Barral and Zoubin Ghahramani and Raia Hadsell and Yossi Matias and D. Sculley and Slav Petrov and Noah Fiedel and Noam Shazeer and Oriol Vinyals and Jeff Dean and Demis Hassabis and Koray Kavukcuoglu and Clement Farabet and Elena Buchatskaya and Jean-Baptiste Alayrac and Rohan Anil and Dmitry and Lepikhin and Sebastian Borgeaud and Olivier Bachem and Armand Joulin and Alek Andreev and Cassidy Hardin and Robert Dadashi and Léonard Hussenot},
      year={2025},
      eprint={2503.19786},
      archivePrefix={arXiv},
      primaryClass={cs.CL},
      url={https://arxiv.org/abs/2503.19786}, 
}

@misc{qwen3,
      title={Qwen3 Technical Report}, 
      author={An Yang and Anfeng Li and Baosong Yang and Beichen Zhang and Binyuan Hui and Bo Zheng and Bowen Yu and Chang Gao and Chengen Huang and Chenxu Lv and Chujie Zheng and Dayiheng Liu and Fan Zhou and Fei Huang and Feng Hu and Hao Ge and Haoran Wei and Huan Lin and Jialong Tang and Jian Yang and Jianhong Tu and Jianwei Zhang and Jianxin Yang and Jiaxi Yang and Jing Zhou and Jingren Zhou and Junyang Lin and Kai Dang and Keqin Bao and Kexin Yang and Le Yu and Lianghao Deng and Mei Li and Mingfeng Xue and Mingze Li and Pei Zhang and Peng Wang and Qin Zhu and Rui Men and Ruize Gao and Shixuan Liu and Shuang Luo and Tianhao Li and Tianyi Tang and Wenbiao Yin and Xingzhang Ren and Xinyu Wang and Xinyu Zhang and Xuancheng Ren and Yang Fan and Yang Su and Yichang Zhang and Yinger Zhang and Yu Wan and Yuqiong Liu and Zekun Wang and Zeyu Cui and Zhenru Zhang and Zhipeng Zhou and Zihan Qiu},
      year={2025},
      eprint={2505.09388},
      archivePrefix={arXiv},
      primaryClass={cs.CL},
      url={https://arxiv.org/abs/2505.09388}, 
}

@inproceedings{rareagents,
  title     = {RareAgents: Autonomous Multi-disciplinary Team for Rare Disease Diagnosis and Treatment},
  author    = {Xuanzhong Chen and Ye Jin and Xiaohao Mao and Lun Wang and Shuyang Zhang and Ting Chen},
  booktitle = {Proceedings of the AAAI Conference on Artificial Intelligence},
  year      = {2026},
  url       = {https://ojs.aaai.org/index.php/AAAI/article/view/36969/40931}
}

@ARTICLE{icd,
  title     = "International classification of diseases, 10th edition, clinical
               modification and procedure coding system: descriptive overview
               of the next generation {HIPAA} code sets",
  author    = "Steindel, Steven J",
  journal   = "J. Am. Med. Inform. Assoc.",
  publisher = "Oxford University Press (OUP)",
  volume    =  17,
  number    =  3,
  pages     = "274--282",
  month     =  may,
  year      =  2010,
  language  = "en"
}

@article{atc,
    doi = {10.1371/journal.pone.0035254},
    author = {Chen, Lei AND Zeng, Wei-Ming AND Cai, Yu-Dong AND Feng, Kai-Yan AND Chou, Kuo-Chen},
    journal = {PLOS ONE},
    publisher = {Public Library of Science},
    title = {Predicting Anatomical Therapeutic Chemical (ATC) Classification of Drugs by Integrating Chemical-Chemical Interactions and Similarities},
    year = {2012},
    month = {04},
    volume = {7},
    url = {https://doi.org/10.1371/journal.pone.0035254},
    pages = {1-7},
    number = {4},

}

@ARTICLE{medi,
  title     = "Development and evaluation of an ensemble resource linking
               medications to their indications",
  author    = "Wei, Wei-Qi and Cronin, Robert M and Xu, Hua and Lasko, Thomas A
               and Bastarache, Lisa and Denny, Joshua C",
  journal   = "J. Am. Med. Inform. Assoc.",
  publisher = "Oxford University Press (OUP)",
  volume    =  20,
  number    =  5,
  pages     = "954--961",
  month     =  sep,
  year      =  2013,
  language  = "en"
}

@ARTICLE{Tatonetti2012-af,
  title     = "Data-driven prediction of drug effects and interactions",
  author    = "Tatonetti, Nicholas P and Ye, Patrick P and Daneshjou, Roxana
               and Altman, Russ B",
  journal   = "Sci. Transl. Med.",
  publisher = "American Association for the Advancement of Science (AAAS)",
  volume    =  4,
  number    =  125,
  pages     = "125ra31",
  month     =  mar,
  year      =  2012,
  language  = "en"
}

@ARTICLE{Kass-Hout2016-vx,
  title     = "{OpenFDA}: an innovative platform providing access to a wealth
               of {FDA's} publicly available data",
  author    = "Kass-Hout, Taha A and Xu, Zhiheng and Mohebbi, Matthew and
               Nelsen, Hans and Baker, Adam and Levine, Jonathan and Johanson,
               Elaine and Bright, Roselie A",
  journal   = "J. Am. Med. Inform. Assoc.",
  publisher = "Oxford University Press (OUP)",
  volume    =  23,
  number    =  3,
  pages     = "596--600",
  month     =  may,
  year      =  2016,
  copyright = "http://creativecommons.org/licenses/by-nc/4.0/",
  language  = "en"
}

@ARTICLE{Ali2023-mh,
  title     = "Deep learning for medication recommendation: A systematic survey",
  author    = "Ali, Zafar and Huang, Yi and Ullah, Irfan and Feng, Junlan and
               Deng, Chao and Thierry, Nimbeshaho and Khan, Asad and Jan, Asim
               Ullah and Shen, Xiaoli and Rui, Wu and Qi, Guilin",
  journal   = "Data Intell.",
  publisher = "China Science Publishing \& Media Ltd.",
  volume    =  5,
  number    =  2,
  pages     = "303--354",
  month     =  may,
  year      =  2023,
  copyright = "https://creativecommons.org/licenses/by/4.0/",
  language  = "en"
}

@article{Shang_Xiao_Ma_Li_Sun_2019, title={GAMENet: Graph Augmented MEmory Networks for Recommending Medication Combination}, volume={33}, url={https://ojs.aaai.org/index.php/AAAI/article/view/3905}, DOI={10.1609/aaai.v33i01.33011126}, abstractNote={&amp;lt;p&amp;gt;Recent progress in deep learning is revolutionizing the healthcare domain including providing solutions to medication recommendations, especially recommending medication combination for patients with complex health conditions. Existing approaches either do not customize based on patient health history, or ignore existing knowledge on drug-drug interactions (DDI) that might lead to adverse outcomes. To fill this gap, we propose the Graph Augmented Memory Networks (GAMENet), which integrates the drug-drug interactions knowledge graph by a memory module implemented as a graph convolutional networks, and models longitudinal patient records as the query. It is trained end-to-end to provide safe and personalized recommendation of medication combination. We demonstrate the effectiveness and safety of GAMENet by comparing with several state-of-the-art methods on real EHR data. GAMENet outperformed all baselines in all effectiveness measures, and also achieved 3&amp;lt;em&amp;gt;.&amp;lt;/em&amp;gt;60% DDI rate reduction from existing EHR data.&amp;lt;/p&amp;gt;}, number={01}, journal={Proceedings of the AAAI Conference on Artificial Intelligence}, author={Shang, Junyuan and Xiao, Cao and Ma, Tengfei and Li, Hongyan and Sun, Jimeng}, year={2019}, month={Jul.}, pages={1126–1133} }

@inproceedings{ijcai2021p514,
  title     = {SafeDrug: Dual Molecular Graph Encoders for Recommending Effective and Safe Drug Combinations},
  author    = {Yang, Chaoqi and Xiao, Cao and Ma, Fenglong and Glass, Lucas and Sun, Jimeng},
  booktitle = {Proceedings of the Thirtieth International Joint Conference on
               Artificial Intelligence, {IJCAI-21}},
  publisher = {International Joint Conferences on Artificial Intelligence Organization},
  editor    = {Zhi-Hua Zhou},
  pages     = {3735--3741},
  year      = {2021},
  month     = {8},
  note      = {Main Track},
  doi       = {10.24963/ijcai.2021/514},
  url       = {https://doi.org/10.24963/ijcai.2021/514},
}

@inproceedings{cognet,
author = {Wu, Rui and Qiu, Zhaopeng and Jiang, Jiacheng and Qi, Guilin and Wu, Xian},
title = {Conditional Generation Net for Medication Recommendation},
year = {2022},
isbn = {9781450390965},
publisher = {Association for Computing Machinery},
address = {New York, NY, USA},
url = {https://doi.org/10.1145/3485447.3511936},
doi = {10.1145/3485447.3511936},
booktitle = {Proceedings of the ACM Web Conference 2022},
pages = {935–945},
numpages = {11},
location = {Virtual Event, Lyon, France},
series = {WWW '22}
}

@inproceedings{molerec,
author = {Yang, Nianzu and Zeng, Kaipeng and Wu, Qitian and Yan, Junchi},
title = {MoleRec: Combinatorial Drug Recommendation with Substructure-Aware Molecular Representation Learning},
year = {2023},
isbn = {9781450394161},
publisher = {Association for Computing Machinery},
address = {New York, NY, USA},
url = {https://doi.org/10.1145/3543507.3583872},
doi = {10.1145/3543507.3583872},
booktitle = {Proceedings of the ACM Web Conference 2023},
pages = {4075–4085},
numpages = {11},
location = {Austin, TX, USA},
series = {WWW '23}
}

@inproceedings{raremed,
author = {Zhao, Zihao and Jing, Yi and Feng, Fuli and Wu, Jiancan and Gao, Chongming and He, Xiangnan},
title = {Leave No Patient Behind: Enhancing Medication Recommendation for Rare Disease Patients},
year = {2024},
isbn = {9798400704314},
publisher = {Association for Computing Machinery},
address = {New York, NY, USA},
url = {https://doi.org/10.1145/3626772.3657785},
doi = {10.1145/3626772.3657785},
booktitle = {Proceedings of the 47th International ACM SIGIR Conference on Research and Development in Information Retrieval},
pages = {533–542},
numpages = {10},
location = {Washington DC, USA},
series = {SIGIR '24}
}

@inproceedings{ultramedical,
 author = {Zhang, Kaiyan and Zeng, Sihang and Hua, Ermo and Ding, Ning and Chen, Zhang-Ren and Ma, Zhiyuan and Li, Haoxin and Cui, Ganqu and Qi, Biqing and Zhu, Xuekai and Lv, Xingtai and Hu, Jin-Fang and Liu, Zhiyuan and Zhou, Bowen},
 booktitle = {Advances in Neural Information Processing Systems},
 doi = {10.52202/079017-0819},
 editor = {A. Globerson and L. Mackey and D. Belgrave and A. Fan and U. Paquet and J. Tomczak and C. Zhang},
 pages = {26045--26081},
 publisher = {Curran Associates, Inc.},
 title = {UltraMedical: Building Specialized Generalists in Biomedicine},
 url = {https://proceedings.neurips.cc/paper_files/paper/2024/file/2dfc26ce9039f00eee4aba0c54931e46-Paper-Datasets_and_Benchmarks_Track.pdf},
 volume = {37},
 year = {2024}
}

@misc{OpenBioLLMs,
  author = {Ankit Pal, Malaikannan Sankarasubbu},
  title = {OpenBioLLMs: Advancing Open-Source Large Language Models for Healthcare and Life Sciences},
  year = {2024},
  publisher = {Hugging Face},
  journal = {Hugging Face repository},
  howpublished = {\url{https://huggingface.co/aaditya/OpenBioLLM-Llama3-70B}}
}

@misc{chen2024huatuogpto1medicalcomplexreasoning,
      title={HuatuoGPT-o1, Towards Medical Complex Reasoning with LLMs}, 
      author={Junying Chen and Zhenyang Cai and Ke Ji and Xidong Wang and Wanlong Liu and Rongsheng Wang and Jianye Hou and Benyou Wang},
      year={2024},
      eprint={2412.18925},
      archivePrefix={arXiv},
      primaryClass={cs.CL},
      url={https://arxiv.org/abs/2412.18925}, 
}

@misc{garciagasulla2025aloefamilyrecipeopen,
      title={The Aloe Family Recipe for Open and Specialized Healthcare LLMs}, 
      author={Dario Garcia-Gasulla and Jordi Bayarri-Planas and Ashwin Kumar Gururajan and Enrique Lopez-Cuena and Adrian Tormos and Daniel Hinjos and Pablo Bernabeu-Perez and Anna Arias-Duart and Pablo Agustin Martin-Torres and Marta Gonzalez-Mallo and Sergio Alvarez-Napagao and Eduard Ayguadé-Parra and Ulises Cortés},
      year={2025},
      eprint={2505.04388},
      archivePrefix={arXiv},
      primaryClass={cs.CL},
      url={https://arxiv.org/abs/2505.04388}, 
}

@article{hager2024evaluation,
  title={Evaluation and mitigation of the limitations of large language models in clinical decision-making},
  author={Hager, Paul and Jungmann, Friederike and Holland, Robbie and Bhagat, Kunal and Hubrecht, Inga and Knauer, Manuel and Vielhauer, Jakob and Makowski, Marcus and Braren, Rickmer and Kaissis, Georgios and Rueckert, Daniel},
  journal={Nature Medicine},
  volume={30},
  pages={2613--2622},
  year={2024},
  doi={10.1038/s41591-024-03097-1}
}

@article{asgari2025framework,
  title={A framework to assess clinical safety and hallucination rates of LLMs for medical text summarisation},
  author={Asgari, Elham and Monta{\~n}a-Brown, Nina and Dubois, Magda and Khalil, Saleh and Balloch, Jasmine and Au Yeung, Joshua and Pimenta, Dominic},
  journal={npj Digital Medicine},
  volume={8},
  pages={274},
  year={2025},
  doi={10.1038/s41746-025-01670-7}
}

@article{farrag2026evaluating,
  title={Evaluating large language models for pharmacotherapy simulations: a mixed-methods study},
  author={Farrag, Ahmed N. and El-Zeiny, Amany and Ali, Amani M.},
  journal={npj Digital Medicine},
  volume={9},
  pages={355},
  year={2026},
  doi={10.1038/s41746-026-02626-1}
}

@misc{zhao2025finegrainedalignmentlargelanguage,
      title={Fine-grained Alignment of Large Language Models for General Medication Recommendation without Overprescription}, 
      author={Zihao Zhao and Chenxiao Fan and Junlong Liu and Zheng Wang and Xiangnan He and Chongming Gao and Juan Li and Fuli Feng},
      year={2025},
      eprint={2503.03687},
      archivePrefix={arXiv},
      primaryClass={cs.IR},
      url={https://arxiv.org/abs/2503.03687}, 
}

@misc{fan2026finegrainedlistwisealignmentgenerative,
      title={Fine-grained List-wise Alignment for Generative Medication Recommendation}, 
      author={Chenxiao Fan and Chongming Gao and Wentao Shi and Yaxin Gong and Zihao Zhao and Fuli Feng},
      year={2026},
      eprint={2505.20218},
      archivePrefix={arXiv},
      primaryClass={cs.LG},
      url={https://arxiv.org/abs/2505.20218}, 
}

@inproceedings{medagents,
    title = "{M}ed{A}gents: Large Language Models as Collaborators for Zero-shot Medical Reasoning",
    author = "Tang, Xiangru  and
      Zou, Anni  and
      Zhang, Zhuosheng  and
      Li, Ziming  and
      Zhao, Yilun  and
      Zhang, Xingyao  and
      Cohan, Arman  and
      Gerstein, Mark",
    editor = "Ku, Lun-Wei  and
      Martins, Andre  and
      Srikumar, Vivek",
    booktitle = "Findings of the Association for Computational Linguistics: ACL 2024",
    month = aug,
    year = "2024",
    address = "Bangkok, Thailand",
    publisher = "Association for Computational Linguistics",
    url = "https://aclanthology.org/2024.findings-acl.33/",
    doi = "10.18653/v1/2024.findings-acl.33",
    pages = "599--621"
}

@inproceedings{mdagents,
 author = {Kim, Yubin and Park, Chanwoo and Jeong, Hyewon and Chan, Yik Siu and Xu, Xuhai and McDuff, Daniel and Lee, Hyeonhoon and Ghassemi, Marzyeh and Breazeal, Cynthia and Park, Hae Won},
 booktitle = {Advances in Neural Information Processing Systems},
 doi = {10.52202/079017-2522},
 editor = {A. Globerson and L. Mackey and D. Belgrave and A. Fan and U. Paquet and J. Tomczak and C. Zhang},
 pages = {79410--79452},
 publisher = {Curran Associates, Inc.},
 title = {MDAgents: An Adaptive Collaboration of LLMs for Medical Decision-Making},
 url = {https://proceedings.neurips.cc/paper_files/paper/2024/file/90d1fc07f46e31387978b88e7e057a31-Paper-Conference.pdf},
 volume = {37},
 year = {2024}
}

@ARTICLE{Zhang2026-li,
  title     = "Knowledge-enhanced Explainable {HyperGraph} Convolution Network
               for medication recommendation",
  author    = "Zhang, Zihan and Liu, Hongzhi and Guo, Xiaoshuang and Sun,
               Tianqi and Wu, Zhonghai",
  journal   = "Proc. Conf. AAAI Artif. Intell.",
  publisher = "Association for the Advancement of Artificial Intelligence
               (AAAI)",
  volume    =  40,
  number    =  19,
  pages     = "16424--16432",
  month     =  mar,
  year      =  2026
}

@article{mimiciii,
  author = {Johnson, Alistair and Pollard, Tom and Mark, Roger},
  title = {{MIMIC-III Clinical Database}},
  journal = {{PhysioNet}},
  year = {2016},
  month = sep,
  note = {Version 1.4},
  doi = {10.13026/C2XW26},
  url = {https://doi.org/10.13026/C2XW26}
}

@article{mimiciv,
  author = {Johnson, Alistair and Bulgarelli, Lucas and Pollard, Tom and Gow, Brian and Moody, Benjamin and Horng, Steven and Celi, Leo Anthony and Mark, Roger},
  title = {{MIMIC-IV}},
  journal = {{PhysioNet}},
  year = {2024},
  month = oct,
  note = {Version 3.1},
  doi = {10.13026/kpb9-mt58},
  url = {https://doi.org/10.13026/kpb9-mt58}
}

@article{xu2022survey,
  title={A survey of deep learning for electronic health records},
  author={Xu, Jiabao and Xi, Xuefeng and Chen, Jie and Sheng, Victor S and Ma, Jieming and Cui, Zhiming},
  journal={Applied Sciences},
  volume={12},
  number={22},
  pages={11709},
  year={2022},
  publisher={MDPI}
}

@book{who2026atc,
  author    = {WHOCC, WHO Collaborating Centre for Drug Statistics Methodology},
  title     = {Guidelines for {ATC} classification and {DDD} assignment},
  address   = {Oslo},
  year      = {2026}
}

@misc{li2024mmedagentlearningusemedical,
      title={MMedAgent: Learning to Use Medical Tools with Multi-modal Agent}, 
      author={Binxu Li and Tiankai Yan and Yuanting Pan and Jie Luo and Ruiyang Ji and Jiayuan Ding and Zhe Xu and Shilong Liu and Haoyu Dong and Zihao Lin and Yixin Wang},
      year={2024},
      eprint={2407.02483},
      archivePrefix={arXiv},
      primaryClass={cs.CL},
      url={https://arxiv.org/abs/2407.02483}, 
}

\clearpage

\appendix
\section{Details About Clinical Domain Knowledge }
\label{app:resources}

\subsection{Construction of Medical Ontologies}
\label{app: med_onto}

We organize diagnosis and medication concepts into hierarchical ontologies to align EHR codes with external clinical knowledge. Diagnosis codes follow the ICD taxonomy based on CCS hierarchy, and medication codes follow the ATC taxonomy. Each code is assigned to a unique path from a broad category to a fine-grained leaf code. This path records the semantic lineage of the code and provides a consistent representation for evidence alignment, medication vocabulary mapping, and safety evaluation.

\paragraph{Diagnosis.} Diagnostic concepts are organized using the CCS framework developed by the Agency for Healthcare Research and Quality (AHRQ), which groups ICD-9-CM and ICD-10-CM diagnosis codes into clinically coherent categories at multiple levels of granularity. We construct a five-level hierarchy from CCS major categories down to decimal-level ICD codes, ensuring that each ICD code maps to a unique hierarchical path.

\paragraph{Medication.}
Medication concepts are organized using the ATC classification system, which defines a five-level hierarchy from anatomical main groups down to chemical substances and encodes pharmacological relationships among drugs.

Per-level concept counts for both taxonomies are reported in Table~\ref{tab:hierarchy_stat}.

\begin{table*}[t]
    \small
    \centering
    \caption{Hierarchical taxonomy of diagnosis and medication concepts used in this study. Each level contains two attributes: \emph{Name}, which specifies the semantic grouping at that level, and \emph{\# Concepts}, the number of unique codes at that level. Diagnosis categories are derived from ICD-CM codes mapped through the CCS diagnostic hierarchy, and medications follow the five-level ATC ontology. For ICD-CM levels, the \emph{\# Concepts} is reported as the number of ICD-9-CM and ICD-10-CM codes shown in parentheses, respectively.}
    \label{tab:hierarchy_stat}
    \begin{tabular}{c|c|cc}
    \toprule
    \textbf{Level} 
    & \textbf{Description} 
    & \textbf{Diagnosis} 
    & \textbf{Medication} \\
    \midrule
    \multirow{2}{*}{\textbf{Level 1}} 
    & Name & Major category & Anatomical main group \\
    & \# Concepts & 18 & 14 \\
    \midrule
    \multirow{2}{*}{\textbf{Level 2}} 
    & Name & Subcategory & Therapeutic subgroup \\
    & \# Concepts & 34 & 90 \\
    \midrule
    \multirow{2}{*}{\textbf{Level 3}} 
    & Name & CCS code & Pharmacological subgroup \\
    & \# Concepts & 265 & 248 \\
    \midrule
    \multirow{2}{*}{\textbf{Level 4}} 
    & Name & Pre-decimal ICD-CM code & Chemical subgroup \\
    & \# Concepts & (937, 1278) & 840 \\
    \midrule
    \multirow{2}{*}{\textbf{Level 5}} 
    & Name & Full ICD-CM code & Chemical substance \\
    & \# Concepts & (13193, 73066) & 5495 \\
    \bottomrule
    \end{tabular}
\end{table*}

\subsection{Medication Indication}
\label{app: medi}

MEDI is an ensemble medication indication resource for primary and secondary uses of EHR data.  MEDI was created based on multiple commonly used medication resources (RxNorm, MedlinePlus, SIDER 4.1, Mayo Clinic, WebMD, and Wikipedia ) and by leveraging both ontology and NLP techniques. The current release of MEDI contains 3,031 medications and 186,064 indications, including both ICD-9-CM and ICD-10-CM indication pairs.

\subsection{Drug-Drug Interactions}
\label{app: ddi}

DDIs describe medication pairs with known co-prescription risks. We construct DDI resources from the TWOSIDES knowledge base~\cite{Tatonetti2012-af}. TWOSIDES contains drug-pair interaction records with source drug identifiers. We map these identifiers to ATC-L4 medication codes and aggregate the mapped records over the ATC-L4 medication vocabulary used in this study. This mapping produces two DDI matrices. The binary DDI matrix records whether an ATC-L4 medication pair has at least one known interaction record in TWOSIDES. If any source drug pair mapped to the same ATC-L4 pair has a recorded interaction, the corresponding binary entry is set to one; otherwise, it is set to zero. This matrix is used to compute binary DDI rates. The weighted DDI matrix records the observed interaction frequency for each ATC-L4 medication pair. If multiple source drug pairs map to the same ATC-L4 pair, we sum their frequencies to obtain one pair-level weight. This matrix is used to compute frequency-weighted DDI rates, assigning larger penalties to medication pairs with more frequently observed interaction records. We symmetrize both matrices because DDI risks are defined over unordered medication pairs. We exclude diagonal entries when computing pairwise DDI rates.

\subsection{Contraindications from openFDA}
\label{app: openfda_processing}

Contraindications specify clinical situations in which a medication may be unsafe because potential harms outweigh expected benefits. The openFDA Drug Labeling dataset provides programmatic access to FDA Structured Product Labeling (SPL) submissions for prescription and over-the-counter drugs~\cite{Kass-Hout2016-vx}. SPL records are organized into clinically meaningful fields, such as indications, contraindications, adverse reactions, and warnings, although field coverage and text format vary across products. The openFDA Drug Labeling data are updated over time as new safety and effectiveness information becomes available.

We use the openFDA Drug Labeling snapshot last updated on 2026-03-04. To construct the ATC-L4 contraindication knowledge database, we first merge all 13 openFDA Drug Labeling files from \url{https://open.fda.gov/data/downloads/}. We retain three fields: \texttt{substance\_name}, \texttt{RxCUI}, and \texttt{contraindications}. We remove records without \texttt{RxCUI} (193{,}958 entries) and records without \texttt{contraindications} (27{,}769 entries). For duplicate single-\texttt{RxCUI} records, we group records by \texttt{RxCUI} and retain the record with the longest contraindication text, measured by $\mathrm{len}(\texttt{contraindications})$. This step reduces single-\texttt{RxCUI} records from 22{,}382 to 2{,}765 unique entries. We then remove all 11{,}152 multi-\texttt{RxCUI} records to avoid ambiguous drug-to-code mappings. Finally, we map the 2{,}765 single-\texttt{RxCUI} records to ATC-L4 codes and retain only successfully mapped entries. This process yields 4{,}434 RxCUI-ATC-L4 pairs, corresponding to 2{,}136 unique RxCUI drugs and 400 unique ATC-L4 codes.

% \begin{table}[t]
%     \small
%     \centering
%     \caption{The summary of the FDA Drug Labeling data after pre-processing and RxCUI-ATC4 mapping. Percentages of missingness are computed over mapped RxCUI-ATC4 pairs (i.e., 4,434).}
%     \begin{tabular}{lcc}
%     \hline
%     \textbf{Measure} & \textbf{Count} & \textbf{Percent (\%)} \\
%     \hline
%     Extracted RxCUI-ATC4 pairs & 4,434 & 100.00 \\
%     Unique RxCUI drugs & 2,136 & -- \\
%     Unique ATC4 drugs & 400 & -- \\
%     \hline
%     Missing \texttt{substance name} & 15 & 0.34 \\
%     Missing \texttt{contraindications} & 36 & 0.81 \\
%     Missing \texttt{drug interactions} & 1,401 & 31.60 \\
%     Missing any of the three fields & 1,446 & 32.61 \\
%     \hline
%     \end{tabular}
%     \label{tab:fda_postmap_summary}
% \end{table}
\section{Expert Construction}
\label{app:expert_construction}

This appendix describes how we construct the expert set used by
\ours{}. The expert taxonomy is derived in two stages. First, we
serialize each patient case using the structured EHR-to-text template
in Figure~\ref{fig:ehr_prompt_template}. We encode the serialized cases into
patient-level text embeddings and sweep the number of clusters over
these embeddings. Second, after selecting the best cluster count, we
inspect the diagnosis composition, ICD-10 chapter distribution, and
representative cases of each cluster to assign clinically interpretable
expert names and chapter mappings. In addition to these cluster-derived
experts, we introduce an always-on universal expert to cover
ICU-common supportive-care medications that are not specific to a
single diagnosis chapter.

\paragraph{EHR-to-text serialization.}
Each structured EHR case is serialized into natural language for
LLM-based methods (Section~\ref{sec:method} and the LLM/Agent baselines in
Section~\ref{sec:experiments}). The serialized input contains the patient
profile, historical visits (diagnoses, procedures, prescribed
medications), and the target visit (diagnoses and procedures only).
Each clinical concept is paired with its original code to preserve
both textual semantics and code-level grounding.
The full template is shown in Fig.~\ref{fig:ehr_prompt_template}.

\begin{figure*}[t]
    \centering
    \begin{tcolorbox}[
        enhanced,
        colback=blue!3,
        colframe=blue!32,
        boxrule=0.6pt,
        arc=2mm,
        left=1.5mm,
        right=1.5mm,
        top=1mm,
        bottom=1mm,
        width=0.96\textwidth,
        title=\textbf{EHR-to-text serialization template},
        fonttitle=\bfseries,
        coltitle=black
    ]
    \scriptsize
    \textbf{Patient profile.}
    The patient's age is [AGE] and gender is [GENDER]. The patient's insurance type
    is [INSURANCE], language is [LANGUAGE], admission type is [ADMISSION TYPE],
    marital status is [MARITAL STATUS], and race is [RACE]. The patient has [T]
    ICU visits.
    
    \medskip
    \textbf{Historical visit.}
    In visit [t], the patient had diagnoses:
    [DIAGNOSIS DESCRIPTION] ([ICD CODE]), \ldots;
    procedures: [PROCEDURE DESCRIPTION] ([PROCEDURE CODE]), \ldots.
    The patient was prescribed drugs:
    [DRUG NAME] ([ATC-L4 CODE]), \ldots.
    
    \medskip
    \textbf{Target visit.}
    In this visit, the patient has diagnoses:
    [DIAGNOSIS DESCRIPTION] ([ICD CODE]), \ldots;
    procedures: [PROCEDURE DESCRIPTION] ([PROCEDURE CODE]), \ldots.
    Then, the patient should be prescribed:
    \end{tcolorbox}
    \caption{\textbf{Structured EHR-to-text serialization template.}
    Textual descriptions are paired with structured codes to preserve both clinical semantics and code-level grounding.}
    \label{fig:ehr_prompt_template}
\end{figure*}

\subsection{Cluster-Derived Specialty Experts}
% \vspace{-0.2cm}

We construct patient-level routing features from diagnosis information
and apply K-means clustering with different values of $K$. Specifically,
we sweep $K \in \{2,3,\ldots,10\}$ and select the value with the best
silhouette score. The sweep identifies $K=7$ as the most separable
solution, achieving a silhouette score of $0.45$. This suggests that the
patient population is best organized into seven coarse clinical groups
under the constructed routing representation.

Figure~\ref{fig:patient_cluster} visualizes the resulting clusters.
The centroid--domain heatmap (Figure~\ref{fig:cluster_centroid}) shows that
each of the seven clusters concentrates its mass on a single clinical domain,
with off-diagonal mass remaining low; this one-cluster-per-domain structure
motivates a one-to-one cluster-to-expert assignment.
The PCA projection of single-domain patients (Figure~\ref{fig:cluster_pca})
confirms that patients with one dominant clinical domain separate visibly along
the first two principal components (53.9\% + 18.5\% explained variance), with
obstetrics/perinatal forming a clearly distinct branch and the remaining six
domains spread along the second axis.

\begin{figure}[t]
    \centering
    \begin{subfigure}[t]{\linewidth}
        \centering
        \includegraphics[width=\linewidth]{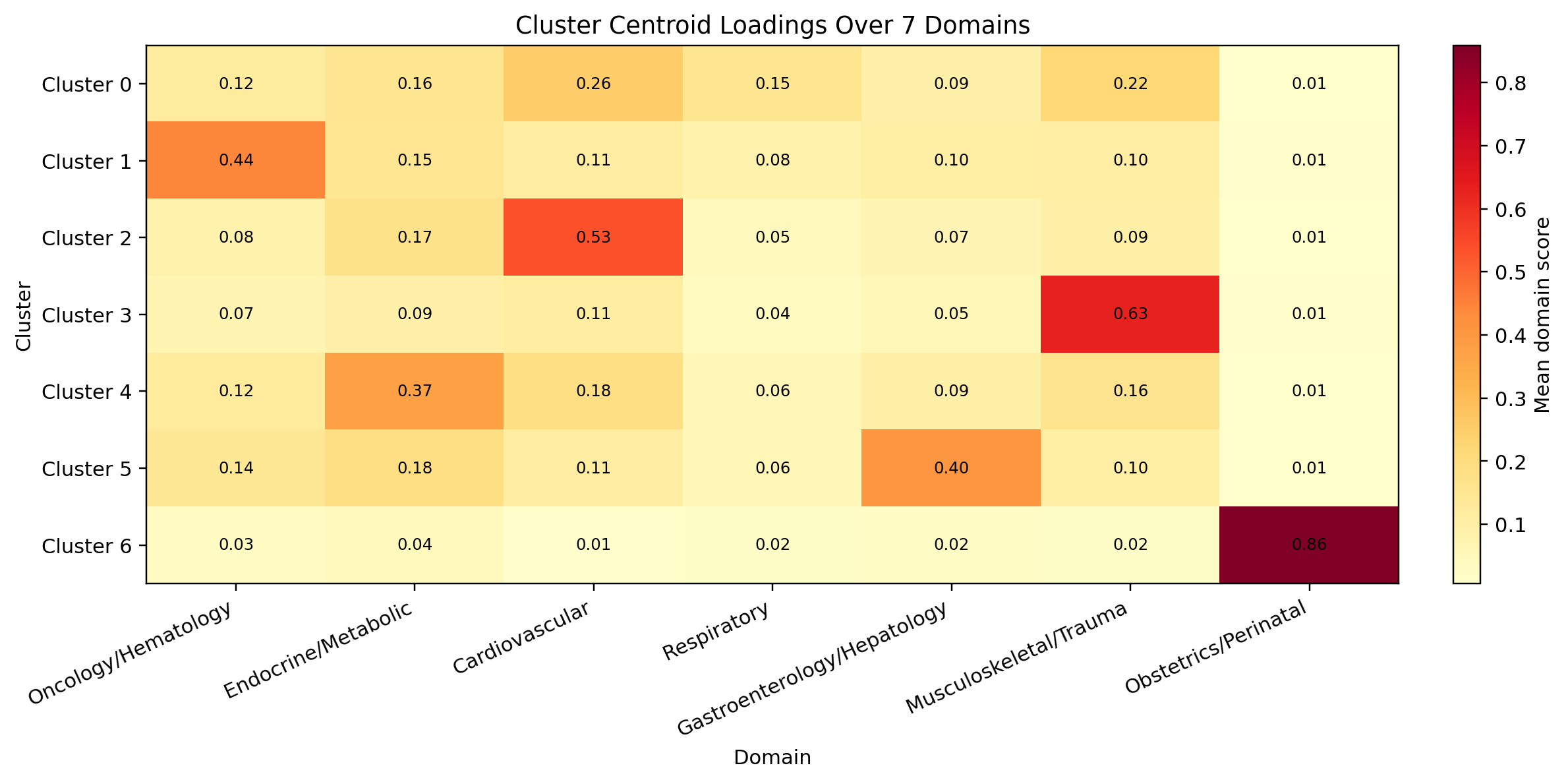}
        \caption{Cluster centroid loadings over the seven derived clinical
        domains. Each cluster concentrates on a single domain.}
        \label{fig:cluster_centroid}
    \end{subfigure}
    \\[0.5em]
    \begin{subfigure}[t]{\linewidth}
        \centering
        \includegraphics[width=\linewidth]{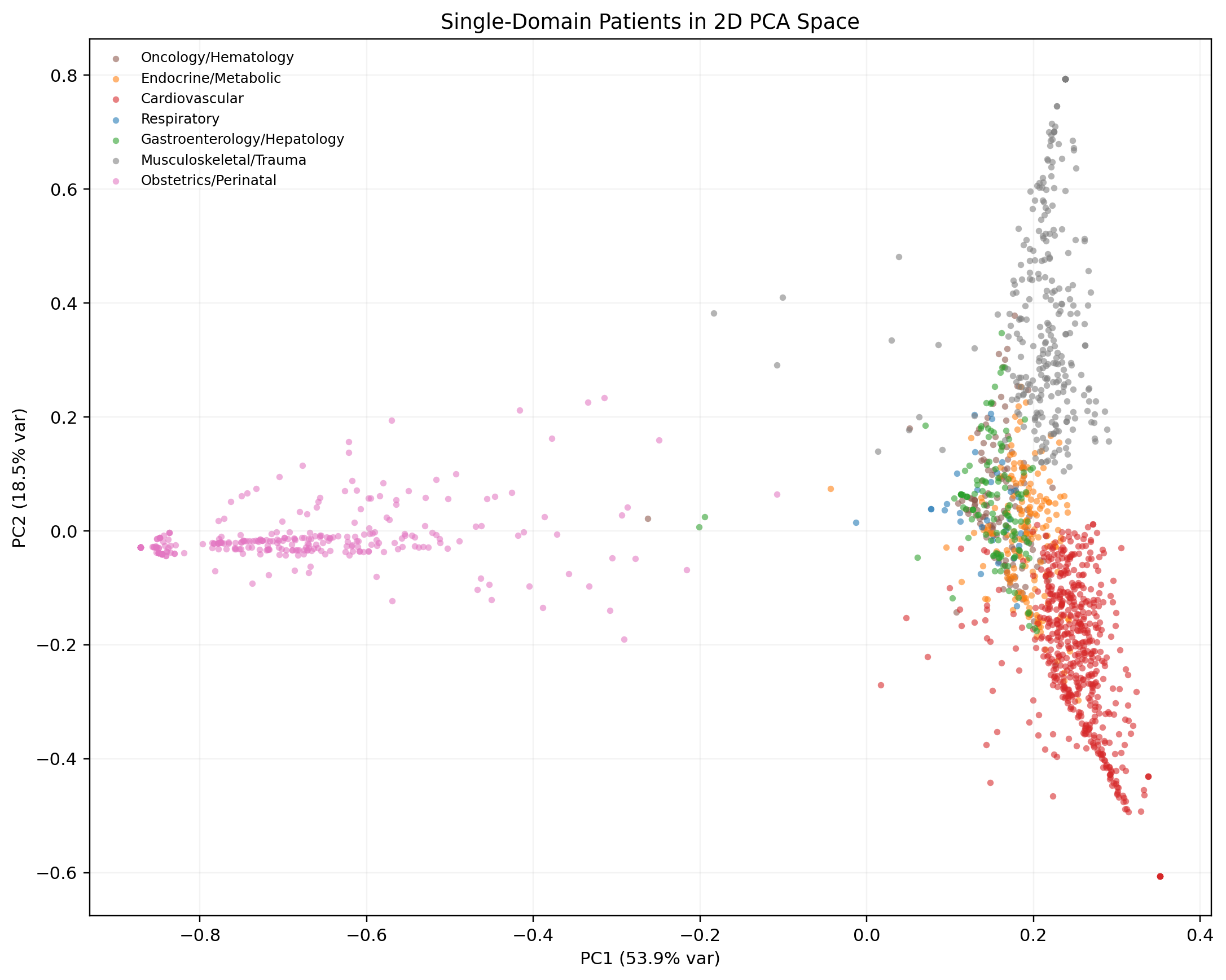}
        \caption{Single-domain patients projected onto the first two
        principal components of the routing feature space, colored by
        dominant clinical domain.}
        \label{fig:cluster_pca}
    \end{subfigure}
    \caption{\textbf{Cluster--domain structure of the routing feature space.}
    The centroid heatmap (top) and PCA projection (bottom) jointly justify
    the $K=7$ partition and the cluster-to-expert mapping in
    Table~\ref{tab:domain_taxonomy}.}
    \label{fig:patient_cluster}
\end{figure}

We then examine each cluster's dominant diagnosis
chapters, diagnosis descriptions, and representative cases. Based on
this inspection, we assign each cluster a clinically interpretable
expert name and map it to the corresponding ICD-10 chapter group.
Table~\ref{tab:domain_taxonomy} summarizes the resulting expert
taxonomy.

\begin{table}[h]
\small
\centering
\begin{tabular}{@{}ll@{}}
\toprule
Expert domain & ICD-10 chapters \\
\midrule
Oncology/Hematology     & II, III \\
Endocrine/Metabolic     & IV \\
Cardiovascular          & IX \\
Respiratory             & X \\
Gastroenterology/Hepatology           & XI \\
Musculoskeletal/Trauma  & XII, XIII, XIX, XX \\
Obstetrics/Perinatal    & XV, XVI, XVII \\
\bottomrule
\end{tabular}
\caption{Cluster-derived specialty experts and their ICD-10 chapter
mappings. The expert names and chapter mappings are curated after
selecting $K=7$ from the clustering sweep.}
\label{tab:domain_taxonomy}
\end{table}
\section{Dataset Statistics}
\label{app:data}
\begin{table}[h]
    \centering
    \footnotesize
    \setlength{\tabcolsep}{5pt}
    \caption{Dataset statistics after preprocessing.}
    \label{tab:data_stats}
    \begin{tabular}{lccc}
    \toprule
    \textbf{Dataset}
    & \textbf{\#Patients}
    & $\boldsymbol{|\mathcal{V}_{\mathrm{med}}|}$
    & \textbf{Avg. \#Target Meds} \\
    \midrule
    MIMIC-III & 901  & 185 & 5.36 \\
    MIMIC-IV  & 1586 & 274 & 9.17 \\
    \bottomrule
    \end{tabular}
\end{table}
\section{Implementation Details}
\label{app:implementation_details}

This appendix provides additional implementation details for LLM inference,
agent-side ablation settings, and prompt templates. All LLM-based methods are
evaluated with the same serialized EHR input format, closed ATC-L4 medication
vocabulary, and evaluation script unless otherwise specified.

\subsection{Inference Settings}
\label{app:inference_settings}

Table~\ref{tab:inference_settings} summarizes the model-level inference
settings used in our experiments. We generate one medication set per case.
Closed-source models are accessed through their official APIs, while
open-source models are served locally via vLLM. For backbone LLMs,
we use temperature 0.0 for deterministic stages (summarization, critique,
and safety verification) and 0.2 for the expert generation stage
to encourage diverse medication proposals across specialists.

\begin{table}[t]
  \centering
  \scriptsize
  \setlength{\tabcolsep}{5pt}
  \caption{\textbf{Inference settings for LLM-based methods.}
  Temp.\ column shows temperatures for deterministic / generation stages;
  medical LLM baselines use only direct prompting (single stage).
  Max model length denotes the maximum context length used for vLLM serving.}
  \label{tab:inference_settings}
  \begin{tabular}{lccc}
  \toprule
  \textbf{Model}
  & \textbf{Access}
  & \textbf{Temp.}
  & \textbf{Max Model Len.} \\
  \midrule
  \multicolumn{4}{c}{\cellcolor{gray!15}\textit{LLM backbones}} \\
  GPT-5.2
  & API
  & 0.0 / 0.2
  & -- \\
  Claude Sonnet-4.6
  & API
  & 0.0 / 0.2
  & -- \\
  Qwen3-32B
  & Local (vLLM)
  & 0.0 / 0.2
  & 8,192 \\
  Gemma3-27B-IT
  & Local (vLLM)
  & 0.0 / 0.2
  & 8,192 \\
  \midrule
  \multicolumn{4}{c}{\cellcolor{gray!15}\textit{Medical LLM baselines}} \\
  UltraMedical-70B
  & Local (vLLM)
  & 0.0
  & 4,096 \\
  Med42-v2-70B
  & Local (vLLM)
  & 0.0
  & 4,096 \\
  Llama3.1-Aloe-Beta-70B
  & Local (vLLM)
  & 0.0
  & 4,096 \\
  HuatuoGPT-o1
  & Local (vLLM)
  & 0.0
  & 4,096 \\
  OpenBioLLM-70B
  & Local (vLLM)
  & 0.0
  & 4,096 \\
  \bottomrule
  \end{tabular}
\end{table}

\section{Evaluation Metrics}
\label{app:evaluation_metrics}

We evaluate each prediction along two axes: \emph{accuracy} and
\emph{safety}. Prediction accuracy is measured using Jaccard, precision,
recall, and F1. Safety is measured using GT-normalized drug--drug interaction
rates and drug--diagnosis contraindication rates. Let $N$ denote the number of
test visits; $\mathcal{M}^{(t)}$ and $\hat{\mathcal{M}}^{(t)}$ denote the
ground-truth and predicted ATC-L4 medication sets for visit $t$; and
$\mathcal{D}^{(t)}$ denote the diagnosis set used for contraindication
screening at that visit.

\subsection{Prediction Accuracy Metrics}

\paragraph{Jaccard.}
We report the sample-averaged Jaccard similarity:
\begin{equation}
\mathrm{Jaccard}
=
\frac{1}{N}\sum_{t=1}^{N}
\frac{\left|\mathcal{M}^{(t)} \cap \hat{\mathcal{M}}^{(t)}\right|}
{\left|\mathcal{M}^{(t)} \cup \hat{\mathcal{M}}^{(t)}\right|}.
\end{equation}

\paragraph{Precision, Recall, and F1.}
We report micro-averaged precision, recall, and F1 by pooling true positives,
false positives, and false negatives across all visits:
\begin{equation}
\mathrm{Precision}
=
\frac{
\sum_{t=1}^{N}
\left|\mathcal{M}^{(t)} \cap \hat{\mathcal{M}}^{(t)}\right|
}{
\sum_{t=1}^{N}
\left|\hat{\mathcal{M}}^{(t)}\right|
},
\end{equation}
\begin{equation}
\mathrm{Recall}
=
\frac{
\sum_{t=1}^{N}
\left|\mathcal{M}^{(t)} \cap \hat{\mathcal{M}}^{(t)}\right|
}{
\sum_{t=1}^{N}
\left|\mathcal{M}^{(t)}\right|
},
\end{equation}
\begin{equation}
\mathrm{F1}
=
\frac{2 \cdot \mathrm{Precision} \cdot \mathrm{Recall}}
{\mathrm{Precision} + \mathrm{Recall}}.
\end{equation}
Micro averaging weights each drug-level decision equally and avoids giving the
same weight to visits with very small and very large medication sets.

\subsection{Prediction Safety Metrics}

\paragraph{GT-normalized DDI rate.}
Drug--drug interactions are represented by a symmetric matrix
$\mathbf{M}_{\mathrm{DDI}}\in[0,1]^{|\mathcal{V}_{\mathrm{med}}|
\times|\mathcal{V}_{\mathrm{med}}|}$ indexed over the ATC-L4 medication vocabulary. We
report two variants: DDI-B uses a binary matrix indicating whether any
documented interaction exists, while DDI-W uses a weighted matrix with
normalized interaction weights.

For each visit $t$, we first compute the predicted interaction burden:
\begin{equation}
I_{\mathrm{DDI}}^{(t)}
=
\sum_{\substack{i<j\\ i,j\in\hat{\mathcal{M}}^{(t)}}}
\mathbf{M}_{\mathrm{DDI}}[i,j].
\end{equation}
We normalize this burden by the number of possible ground-truth drug pairs:
\begin{equation}
Z_{\mathrm{DDI}}^{(t)}
=
\max\!\left(\binom{|\mathcal{M}^{(t)}|}{2}, 1\right).
\end{equation}
The per-visit GT-normalized DDI score is:
\begin{equation}
s_{\mathrm{DDI}}^{(t)}
=
\min\!\left(
\frac{I_{\mathrm{DDI}}^{(t)}}{Z_{\mathrm{DDI}}^{(t)}},
1
\right).
\end{equation}
We report the dataset-level DDI score as:
\begin{equation}
\mathrm{DDI}_{\mathrm{GTnorm}}
=
\frac{1}{N}\sum_{t=1}^{N}s_{\mathrm{DDI}}^{(t)} .
\end{equation}

Unlike the conventional DDI rate, which normalizes by the number of predicted
drug pairs, this metric normalizes the predicted interaction burden by the
expected prescription complexity of the corresponding visit. This prevents large
prediction sets from artificially lowering the DDI score by adding many
non-interacting pairs. The per-visit score is capped at $1$ so that visits with
excessive interaction burden are treated as maximally unsafe without dominating
the dataset average. Lower values indicate safer predictions.

\paragraph{GT-normalized contraindication rate.}
Drug--diagnosis contraindications are represented by a bipartite matrix
$\mathbf{M}_{\mathrm{Contra}}\in[0,1]^{|\mathcal{V}_{\mathrm{med}}|
\times|\mathcal{V}_{\mathrm{diag}}|}$, where $\mathbf{M}_{\mathrm{Contra}}[m,d]$ indicates whether
medication $m$ is contraindicated for diagnosis $d$ under the binary variant, or gives
a normalized contraindication weight under the weighted variant.

For each visit $t$, we first compute the predicted contraindication burden:
\begin{equation}
I_{\mathrm{contra}}^{(t)}
=
\sum_{m\in\hat{\mathcal{M}}^{(t)}}
\sum_{d\in\mathcal{D}^{(t)}}
\mathbf{M}_{\mathrm{Contra}}[m,d].
\end{equation}
We normalize this burden by the ground-truth medication-set size and the number
of active diagnoses:
\begin{equation}
Z_{\mathrm{contra}}^{(t)}
=
\max\!\left(
|\mathcal{M}^{(t)}|\,|\mathcal{D}^{(t)}|,
1
\right).
\end{equation}
The per-visit GT-normalized contraindication score is:
\begin{equation}
s_{\mathrm{contra}}^{(t)}
=
\min\!\left(
\frac{I_{\mathrm{contra}}^{(t)}}{Z_{\mathrm{contra}}^{(t)}},
1
\right).
\end{equation}
We report the dataset-level contraindication score as:
\begin{equation}
\mathrm{Contra}_{\mathrm{GTnorm}}
=
\frac{1}{N}\sum_{t=1}^{N}s_{\mathrm{contra}}^{(t)} .
\end{equation}

Indices that fall outside the matrix support contribute zero to the numerator
under both metrics: drug pairs $(i,j)$ outside
$\mathbf{M}_{\mathrm{DDI}}$ for DDI-B / DDI-W and drug--diagnosis pairs $(m,d)$
outside $\mathbf{M}_{\mathrm{Contra}}$ for Contra-B / Contra-W. This convention
treats safety-knowledge--unknown pairs as safe-by-default, so the reported
DDI / contraindication rates are lower bounds on the true safety burden over
$\mathcal{V}_{\mathrm{med}}$. Lower values indicate safer predictions.
\section{Additional Diagnostic Analysis}
\label{app:diagnostic_analysis}

This appendix complements Section~\ref{sec:diagnostic_analysis} with two
additional diagnostics for the critique stage: the overall action distribution
(Table~\ref{tab:critique_action_app}) and per-expert removal behavior
(Table~\ref{tab:critique_expert_removal_app}).

\begin{table}[t]
\centering
\footnotesize
\setlength{\tabcolsep}{6pt}
\caption{\textbf{Critique action distribution.}
Share\% is the fraction of all critique decisions assigned to each action.
TP\% and FP\% report the composition of retained and removed medications with
respect to the ground-truth medication set.}
\label{tab:critique_action_app}
\begin{tabular}{lrrrr}
\toprule
\textbf{Action} & \textbf{Count} & \textbf{Share\%} & \textbf{TP\%} & \textbf{FP\%} \\
\midrule
\textsc{ret} & 16098 & 70.8 & 37.8 & 62.2 \\
\textsc{rem} & 6625 & 29.2 & 11.7 & 88.3 \\
\bottomrule
\end{tabular}
\end{table}

\begin{table}[t]
\centering
\scriptsize
\setlength{\tabcolsep}{4pt}
\caption{\textbf{Per-expert removal behavior.}
Rm\% denotes the fraction of proposed medications removed by critique. RmFP\%
denotes the false-positive rate among removed medications.}
\label{tab:critique_expert_removal_app}
\begin{tabular}{lrrrr}
\toprule
\textbf{Expert} & \textbf{Proposed} & \textbf{Removed} & \textbf{Rm\%} & \textbf{RmFP\%} \\
\midrule
Universal supportive & 12345 & 822 & 6.7 & 84.5 \\
Endocrine/metabolic & 4194 & 745 & 17.8 & 85.4 \\
Cardiovascular & 4847 & 1128 & 23.3 & 82.7 \\
Gastroenterology/hepatology & 2391 & 602 & 25.2 & 84.6 \\
Musculoskeletal/trauma & 4907 & 1420 & 28.9 & 88.1 \\
Oncology/hematology & 2993 & 920 & 30.7 & 93.2 \\
Obstetrics/perinatal & 2344 & 758 & 32.3 & 96.6 \\
Respiratory & 1308 & 490 & 37.5 & 90.6 \\
\bottomrule
\end{tabular}
\end{table}

Removal behavior differs across experts. The universal supportive-care expert
has the lowest removal rate, consistent with its role in proposing common ICU
medications. Specialty experts have higher removal rates and consistently high
RmFP\%, indicating that critique mostly filters overly broad or weakly
supported specialty-specific proposals rather than indiscriminately pruning
correct ones.
\section{Additional Expert Analysis}
\label{app:expert_analysis}

This appendix complements the leave-one-expert-out analysis in
Section~\ref{sec:expert_analysis} with two further diagnostics:
router activation behavior (Figure~\ref{fig:expert_activation_app}) and
expert-level proposal/retention statistics (Table~\ref{tab:expert_contribution_app}).
Expert abbreviations used throughout this appendix and in
Figure~\ref{fig:expert_subgroup}: SUP = universal supportive, CVD =
cardiovascular, ENDO = endocrine/metabolic, MSK = musculoskeletal/trauma,
ONC = oncology/hematology, OB = obstetrics/perinatal, GI =
gastroenterology/hepatology, RESP = respiratory.

The cardiovascular expert has the highest TP/Ret.\ ratio (51.5\%), suggesting
that specialty experts contribute targeted domain-specific candidates
beyond what broad supportive-care coverage provides; specialty experts that
are activated less frequently (e.g., RESP, OB) still maintain TP/Ret.\ above
35\%, indicating that the router does not over-activate marginal experts.

\begin{figure}[t]
    \centering
    \begin{subfigure}[t]{\linewidth}
        \centering
        \includegraphics[width=\linewidth]{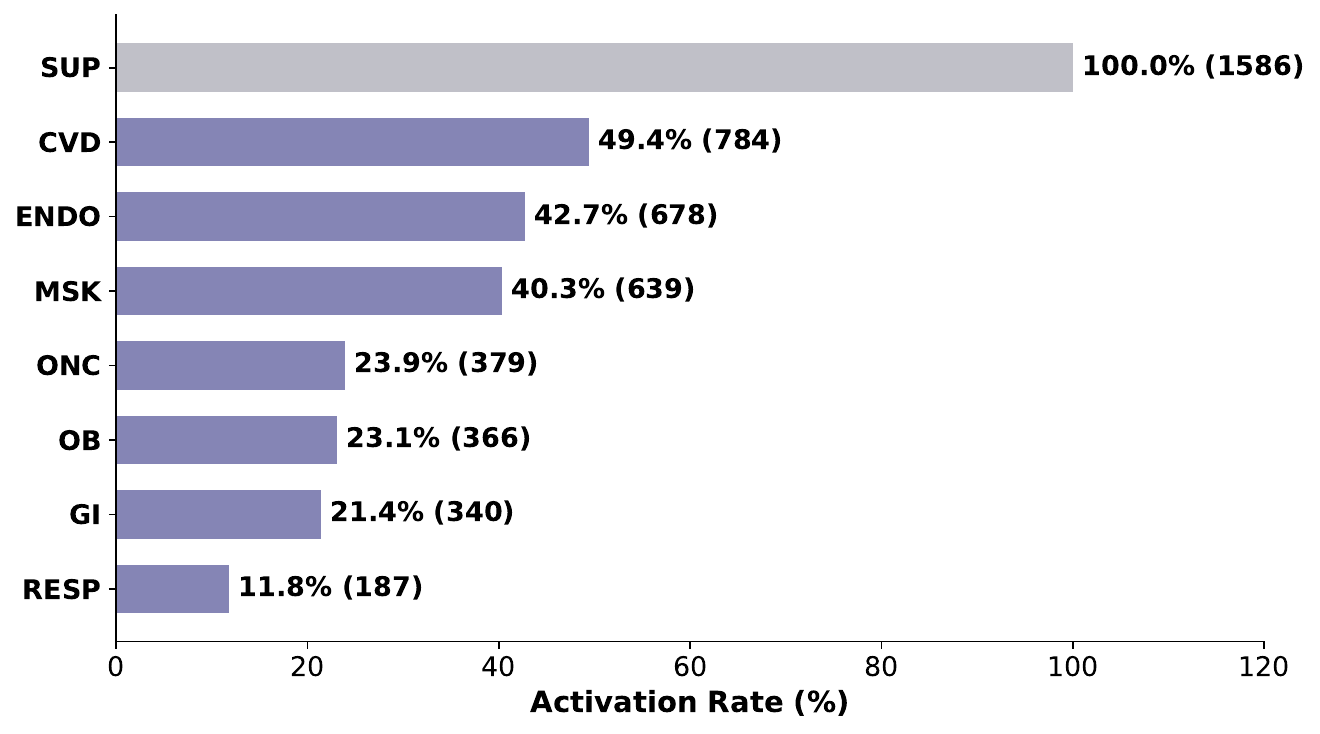}
        \caption{Activation frequency.}
        \label{fig:expert_activation_frequency_app}
    \end{subfigure}
    \\[0.5em]
    \begin{subfigure}[t]{\linewidth}
        \centering
        \includegraphics[width=\linewidth]{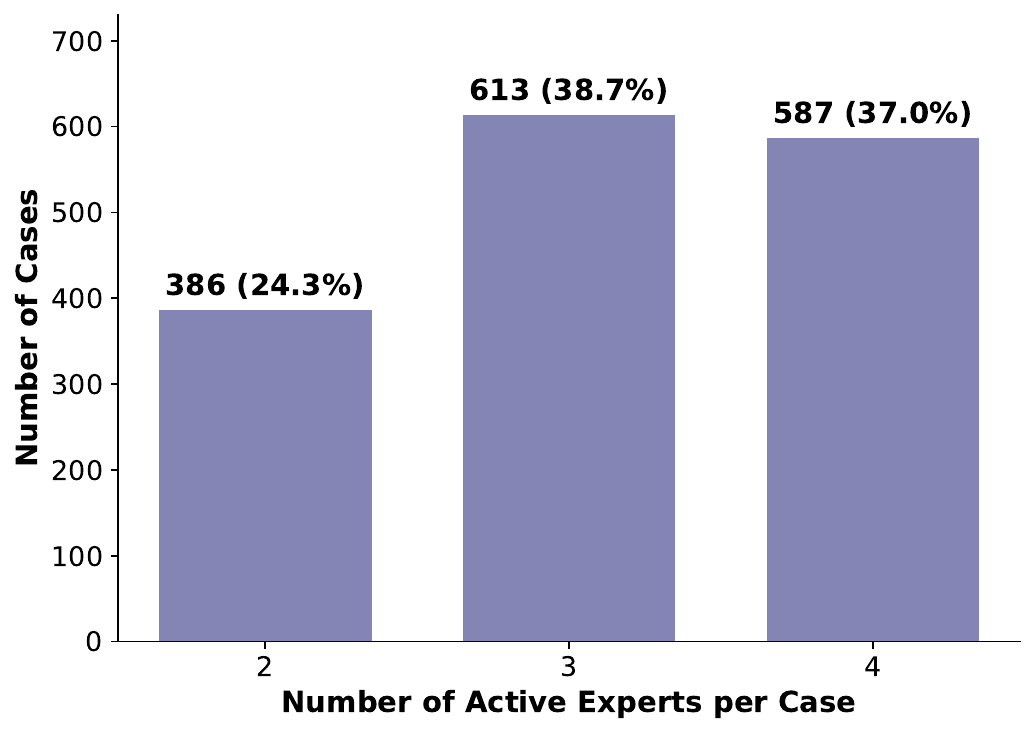}
        \caption{Number of active experts per case.}
        \label{fig:expert_count_distribution_app}
    \end{subfigure}
    \caption{\textbf{Expert activation statistics.}
    The router activates a sparse, case-dependent subset of experts. Expert
abbreviations are defined at the beginning of Appendix~\ref{app:expert_analysis}.}
    \label{fig:expert_activation_app}
\end{figure}

\begin{table*}[t]
    \centering
    \small
    \setlength{\tabcolsep}{5pt}
    \caption{\textbf{Expert contribution statistics.}
    Activated counts selected cases. Proposed and retained count expert-level
    medication candidates before and after critic filtering. TP/Ret. is the
    true-positive density among retained predictions.}
    \label{tab:expert_contribution_app}
    \begin{tabular}{lrrrrrr}
    \toprule
    \textbf{Expert} & \textbf{Activated} & \textbf{Proposed} & \textbf{Retained} & \textbf{Ret.\%} & \textbf{TP} & \textbf{TP/Ret.} \\
    \midrule
    Universal supportive
    & 1586 & 12345 & 11523 & 93.3 & 4573 & 39.7 \\
    Cardiovascular 
    & 832 & 4847 & 3719 & 76.7 & 1915 & 51.5 \\
    Musculoskeletal/trauma 
    & 672 & 4907 & 3487 & 71.1 & 1633 & 46.8 \\
    Endocrine/metabolic 
    & 721 & 4194 & 3449 & 82.2 & 1581 & 45.8 \\
    Gastroenterology/hepatology 
    & 360 & 2391 & 1789 & 74.8 & 796 & 44.5 \\
    Oncology/hematology 
    & 405 & 2993 & 2073 & 69.3 & 874 & 42.2 \\
    Respiratory 
    & 197 & 1308 & 818 & 62.5 & 354 & 43.3 \\
    Obstetrics/perinatal 
    & 393 & 2344 & 1586 & 67.7 & 570 & 35.9 \\
    \bottomrule
    \end{tabular}
\end{table*}
\section{Efficiency Breakdown}
\label{app:efficiency_breakdown}

Table~\ref{tab:efficiency_breakdown} reports the average LLM calls, input
tokens, output tokens, and wall-clock latency per case for each operator in
Algorithm~\ref{alg:saferx}. \textsc{Summarize}, \textsc{Generate}, and
\textsc{Verify} are invoked per activated expert or per flagged item, so
their per-case totals scale with case complexity. \textsc{Route}
(Section~\ref{sec:generation}) and \textsc{FindFlags} (Section~\ref{sec:safety}) are
deterministic and contribute negligible cost.

\begin{table}[t]
\centering
\scriptsize
\setlength{\tabcolsep}{3pt}
\caption{\textbf{Per-phase efficiency breakdown of SafeRx-Agent on MIMIC-IV.}
All numbers are averaged over evaluation cases. \textsc{Route} and \textsc{FindFlags} are
deterministic (no LLM calls).}
\label{tab:efficiency_breakdown}
\begin{tabular}{lrrrr}
\toprule
\textbf{Phase} & \textbf{Calls} & \textbf{In Tokens} & \textbf{Out Tokens} & \textbf{Sec.} \\
\midrule
\multicolumn{5}{c}{\cellcolor{gray!15}\textit{Multi-agent generation}} \\
\textsc{Route}      & 0  & 0  & 0  & $<$0.01 \\
\textsc{Summarize}  & 3.7 & 5,010 & 2,559 & 5.7      \\
\textsc{Generate}   & 3.7 & 11,475 & 2,329 & 13.1      \\
\textsc{Critique}   & 1.0 & 7,775 & 1,100 & 8.9      \\
\midrule
\multicolumn{5}{c}{\cellcolor{gray!15}\textit{Safety verification}} \\
\textsc{FindFlags}  & 0  & 0  & 0  & $<$0.01 \\
\textsc{Verify}     & 1.0 & 587 & 82 & 1.8      \\
\midrule
Total w/o safety filter & 8.4 & 24{,}260 & 5{,}988 & 27.7 \\
Total w/ safety filter  & 9.4 & 24{,}847 & 6{,}070 & 29.5 \\
\bottomrule
\end{tabular}
\end{table}

\FloatBarrier
\section{Prompt Templates}
\label{app:prompts}

This section provides all prompt templates used in our experiments.
Appendix~\ref{app:saferx_prompts} lists the templates for the LLM-based
operators in \ours{}, and Appendix~\ref{app:baseline_prompts} lists the
baseline templates for direct prompting, the general-agent baseline, and the
RareAgents-style baseline.

\subsection{SafeRx-Agent Operator Prompts}
\label{app:saferx_prompts}

Figures~\ref{fig:prompt_summarizer}, \ref{fig:prompt_generator},
\ref{fig:prompt_critique}, and \ref{fig:prompt_safety_verifier} correspond
to \textsc{Summarize}, \textsc{Generate}, \textsc{Critique}, and
\textsc{Verify}, respectively. The non-LLM operators \textsc{Route} and
\textsc{FindFlags} are deterministic and do not require prompt templates.

\newcounter{savedfigcount}
\setcounter{savedfigcount}{\value{figure}}
\setcounter{figure}{14}
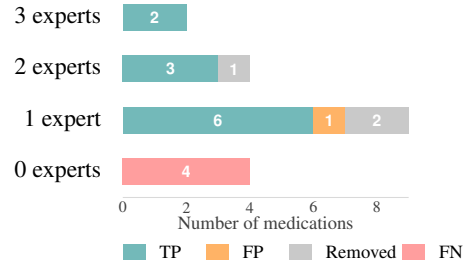
\begin{figure}[!t]
\centering
\begin{tikzpicture}[yscale=0.85]

% Bar parameters
\def\bh{0.42}   % bar half-height
\def\u{0.42}    % width per medication count
\def\ya{2.4}    % y for 3 experts
\def\yb{1.6}    % y for 2 experts
\def\yc{0.8}    % y for 1 expert
\def\yd{0}      % y for 0 experts

% Row labels
\node[anchor=east, font=\small] at (-0.15, \ya) {3 experts};
\node[anchor=east, font=\small] at (-0.15, \yb) {2 experts};
\node[anchor=east, font=\small] at (-0.15, \yc) {1 expert};
\node[anchor=east, font=\small] at (-0.15, \yd) {0 experts};

% 3 experts: 2 TP
\fill[teal!55] (0, \ya-\bh/2) rectangle ({2*\u}, \ya+\bh/2);
\node[font=\sffamily\tiny\bfseries, white] at ({1*\u}, \ya) {2};

% 2 experts: 3 TP + 1 Removed
\fill[teal!55] (0, \yb-\bh/2) rectangle ({3*\u}, \yb+\bh/2);
\fill[gray!42] ({3*\u}, \yb-\bh/2) rectangle ({4*\u}, \yb+\bh/2);
\node[font=\sffamily\tiny\bfseries, white] at ({1.5*\u}, \yb) {3};
\node[font=\sffamily\tiny\bfseries, white] at ({3.5*\u}, \yb) {1};

% 1 expert: 6 TP + 1 FP + 2 Removed
\fill[teal!55] (0, \yc-\bh/2) rectangle ({6*\u}, \yc+\bh/2);
\fill[orange!60] ({6*\u}, \yc-\bh/2) rectangle ({7*\u}, \yc+\bh/2);
\fill[gray!42] ({7*\u}, \yc-\bh/2) rectangle ({9*\u}, \yc+\bh/2);
\node[font=\sffamily\tiny\bfseries, white] at ({3*\u}, \yc) {6};
\node[font=\sffamily\tiny\bfseries, white] at ({6.5*\u}, \yc) {1};
\node[font=\sffamily\tiny\bfseries, white] at ({8*\u}, \yc) {2};

% 0 experts: 4 FN
\fill[red!38] (0, \yd-\bh/2) rectangle ({4*\u}, \yd+\bh/2);
\node[font=\sffamily\tiny\bfseries, white] at ({2*\u}, \yd) {4};

% Axis
\draw[gray!30] (0, -0.4) -- ({9*\u}, -0.4);
\foreach \x in {0,2,4,6,8} {
    \node[font=\tiny, gray!50!black] at ({\x*\u}, -0.55) {\x};
}
\node[font=\scriptsize, gray!50!black] at ({4.5*\u}, -0.8) {Number of medications};

% Legend
\def\ly{-1.25}
\fill[teal!55] (0, \ly-0.12) rectangle (0.3, \ly+0.12);
\node[font=\scriptsize, anchor=west] at (0.35, \ly) {TP};
\fill[orange!60] (1.1, \ly-0.12) rectangle (1.4, \ly+0.12);
\node[font=\scriptsize, anchor=west] at (1.45, \ly) {FP};
\fill[gray!42] (2.2, \ly-0.12) rectangle (2.5, \ly+0.12);
\node[font=\scriptsize, anchor=west] at (2.55, \ly) {Removed};
\fill[red!38] (3.7, \ly-0.12) rectangle (4.0, \ly+0.12);
\node[font=\scriptsize, anchor=west] at (4.05, \ly) {FN};

\end{tikzpicture}
\caption{\textbf{Prediction outcome by expert support} (case study, Appendix~\ref{app:case_analysis}). Medications proposed by more experts are more likely to be true positives. All 5 codes with $\geq$2~expert support that were retained by \textsc{Critique} are correct (TP). The only false positive (A06AB) and all 3 correctly removed codes were single-expert proposals. The 4 FN medications were not proposed by any expert.}
\label{fig:case_support}
\end{figure}
\setcounter{figure}{\value{savedfigcount}}

\begin{figure*}[t]
\begin{tcolorbox}[
    enhanced,
    width=\textwidth,
    colback=blue!3,
    colframe=blue!32,
    title=\ours{} \textsc{Summarize} Prompt Constructor,
    fonttitle=\bfseries,
    coltitle=black,
    fontupper=\footnotesize,
]
\footnotesize
\textbf{System prompt:}

You are the activated specialty module inside a routed inpatient medication
prediction system. The upstream router selected this skill; use the routing
selection as an interpretive lens and not as permission to invent facts not
present in the patient record.

\vspace{0.5em}
\textbf{Input:}
\begin{itemize}[leftmargin=1.2em,itemsep=0.15em,topsep=0.2em]
    \item Skill name and profile.
    \item Available tools.
    \item Summarization playbook (expert-specific).
    \item Patient record.
\end{itemize}

\vspace{0.5em}
\textbf{Task rules:}
\begin{itemize}[leftmargin=1.2em,itemsep=0.15em,topsep=0.2em]
    \item Produce a specialty-aware medication summary by reading the
    patient record through the lens of the activated expert's playbook.

    \item Keep the summary grounded in the record only; do not invent
    diagnoses, procedures, or medications that are not documented.

    \item Highlight domain-specific risks, procedures, care context, and
    prior medication evidence that could affect inpatient medication
    classes.

    \item Do not predict medications at this stage; medication proposals
    are produced downstream by the \textsc{Generate} operator.
\end{itemize}

\vspace{0.5em}
\textbf{Output format:}

Return strict JSON with five fields: \emph{expertise} (the activated expert
identity), \emph{current\_admission} (specialty-relevant active problems,
organ dysfunction, and care context), \emph{medication\_relevant\_history}
(prior indications, procedures/devices, and explicit medication evidence),
\emph{expertise\_focus} (record-grounded facts explaining why this expert is
relevant), and \emph{risks\_to\_watch} (specialty-relevant medication
risks).

\vspace{0.5em}
\textbf{Example playbook (universal supportive expert):}

Focus the summary on factors that drive universal supportive medication
needs:
\begin{itemize}[leftmargin=1.2em,itemsep=0.15em,topsep=0.2em]
    \item \textbf{ICU / critical-care context}: ICU admission, intubation,
    vasopressor use, post-surgical, or otherwise critically ill -- drives
    prophylactic needs.

    \item \textbf{Immobility / VTE risk}: bed-bound, post-operative, trauma, or
    prolonged hospitalization $\rightarrow$ VTE prophylaxis.

    \item \textbf{GI stress / aspiration risk}: NPO status, intubation,
    critical illness, steroid use $\rightarrow$ stress-ulcer prophylaxis,
    bowel management, antiemetics.

    \item \textbf{Pain / sedation needs}: surgical procedures, trauma, burns,
    invasive lines $\rightarrow$ analgesics, sedatives.

    \item \textbf{Electrolyte context}: renal function, diuretic use, critical
    illness $\rightarrow$ electrolyte monitoring and repletion.

    \item \textbf{Nutritional status}: NPO, TPN, tube feeding, prolonged
    admission $\rightarrow$ vitamins, IV fluids, dextrose.

    \item \textbf{Infection / vaccine status}: any hospitalization
    $\rightarrow$ influenza-vaccine consideration; critical illness
    $\rightarrow$ infection prophylaxis.

    \item \textbf{Prior medications}: note supportive medications from prior
    visits that suggest continuation.
\end{itemize}
Do not summarize specialty-specific clinical details (e.g., cardiac rhythm,
tumor staging); those belong to routed specialty experts.
\end{tcolorbox}
\caption{
Prompt template for the \ours{} \textsc{Summarize} operator described in
Section~\ref{sec:generation}. The prompt is dynamically constructed per
activated expert by injecting the expert-specific playbook into the shared
template; the universal supportive expert's playbook is shown as an
example.}
\label{fig:prompt_summarizer}
\end{figure*}

\begin{figure*}[t]
\begin{tcolorbox}[
    enhanced,
    width=\textwidth,
    colback=blue!3,
    colframe=blue!32,
    title=\ours{} \textsc{Generate} Prompt Template,
    fonttitle=\bfseries,
    coltitle=black,
    fontupper=\footnotesize,
]
\textbf{System prompt:}

You are the medication generator for the activated expertise skill. This is a
medication-list prediction task for the current admission; work only from the
expert-specific summary and the patient record. Do not output guideline-only
wish lists -- predict what is plausibly on the medication list.

\vspace{0.5em}
\textbf{Input:}

Expert-specific clinical summary (the JSON produced by \textsc{Summarize});
expert-specific drug-prediction checklist; prior-visit medication evidence;
retrieved indication candidates from MEDI; and optional revision feedback
from a prior generation round.

\vspace{0.5em}
\textbf{Task rules:}
\begin{itemize}[leftmargin=1.2em,itemsep=0.15em,topsep=0.2em]
    \item Predict ATC-L4 medication classes likely present during the current
    admission. Output only 5-character ATC-L4 codes from the closed
    vocabulary; convert any ingredient- or product-level ATC codes to their
    ATC-L4 parent.

    \item Map from clinical scene to plausible medication class, then choose
    the fitting ATC-L4 code; do not return an empty list when the summary
    clearly supports inpatient medication classes.

    \item Follow this evidence priority (strongest first): (i) explicit
    medication evidence, (ii) prior-visit medications, (iii) procedure,
    device, or care-context evidence, (iv) current diagnoses and acute
    organ dysfunction, (v) retrieved indication candidates, (vi) supportive
    inpatient priors when supported by the documented scene.

    \item When a revision block is present, address the prior critique by
    adjusting only the affected codes; do not regenerate the entire list
    from scratch.
\end{itemize}

\vspace{0.5em}
\textbf{Output format:}

Return strict JSON containing the predicted ATC-L4 codes, per-code confidence
scores, input-grounded reasons that cite specific summary fields, any
assumptions that limit certainty, and plausible alternative codes when the
evidence is ambiguous.

\vspace{0.5em}
\textbf{Expert-specific checklist (example: universal supportive expert).}
The checklist gives per-class predict/withhold rules grounded in documented
clinical conditions. For the universal supportive expert it covers four
groups:

\begin{itemize}[leftmargin=1.2em,itemsep=0.15em,topsep=0.2em]
    \item \textbf{Thromboprophylaxis} (heparins, antiplatelets) ---
    predict when VTE prophylaxis is documented, the patient is
    critically ill or post-operative, or antiplatelet therapy is indicated
    by coronary/cerebrovascular disease; withhold when active bleeding,
    HIT, or an anticoagulation contraindication is present.

    \item \textbf{Analgesia} (opioids, acetaminophen, salicylates) ---
    predict for major surgery, severe trauma, documented pain or fever,
    or an active analgesia-sedation protocol; do not predict from mild
    pain, minor procedures, or ICU admission alone.

    \item \textbf{GI prophylaxis} (acid suppression, antiemetics, laxatives)
    --- predict when GI bleeding history, stress-ulcer risk factors
    (mechanical ventilation \emph{with} coagulopathy or shock),
    opioid-induced constipation, or postoperative nausea is documented;
    do not predict from ICU stay alone or in the presence of diarrhea or
    bowel obstruction.

    \item \textbf{Electrolyte and nutrition support} (calcium, potassium,
    dextrose, parenteral nutrition) --- predict when specific electrolyte
    depletion, diuretic use, renal impairment, insulin infusion, or
    enteral/parenteral feeding is documented; do not predict from IV access
    or NPO status alone.
\end{itemize}
\end{tcolorbox}
\caption{
Prompt template for the \ours{} \textsc{Generate} operator described in
Section~\ref{sec:generation}. An expert-specific drug-prediction checklist is
injected at runtime to constrain drug-class prediction to the activated
expert's scope.}
\label{fig:prompt_generator}
\end{figure*}

\begin{figure*}[t]
\begin{tcolorbox}[
    enhanced,
    width=\textwidth,
    colback=blue!3,
    colframe=blue!32,
    title=\ours{} \textsc{Critique} Prompt,
    fonttitle=\bfseries,
    coltitle=black,
    fontupper=\footnotesize,
]
\textbf{System prompt:}

You are the attending physician critique. Multiple specialist experts have
proposed ATC-L4 medication classes for this inpatient admission. Your job is to
produce the final prescription list by conservatively filtering the union of
proposed codes.

\vspace{0.5em}
\textbf{Input:}

Clinical summary; raw patient record; expert proposals with rationales;
prior-visit medication evidence; and the union of proposed ATC-L4 codes.

\vspace{0.5em}
\textbf{Audit rules:}
\begin{itemize}[leftmargin=1.2em,itemsep=0.15em,topsep=0.2em]
    \item Review each proposed code using the clinical summary, raw record,
    expert rationales, prior-medication evidence, and inpatient care context.

    \item Keep a code if it has concrete patient-specific support from any
    expert or record evidence; if it is a prior-visit medication whose
    underlying condition appears ongoing; or if it is a plausible inpatient
    supportive or prophylactic medication.

    \item Remove a code only if it is unsupported by any evidence in the
    summary, raw record, expert proposals, or prior-medication context, and has
    no plausible inpatient reason given the documented conditions, procedures,
    or care context.

    \item Do not remove a code simply because it is outside a specialist's
    scope. Do not remove evidence-grounded prophylactic or supportive
    medications unless there is specific evidence that they are contraindicated
    or absent.

    \item Remove redundant or overlapping codes only when they clearly duplicate
    the same therapeutic intent without added benefit.

    \item When in doubt, keep the code. Do not add codes outside the union.
\end{itemize}

\vspace{0.5em}
\textbf{Output format:}

Return strict JSON containing the final retained ATC-L4 codes, removed codes
with brief removal reasons, an overall rationale, and any missing information
needed for clarification.
\end{tcolorbox}
\caption{
Abbreviated prompt template for the \ours{} \textsc{Critique} operator.}
\label{fig:prompt_critique}
\end{figure*}

\begin{figure*}[t]
\begin{tcolorbox}[
    enhanced,
    width=\textwidth,
    colback=blue!3,
    colframe=blue!32,
    title=\ours{} \textsc{Verify} Prompt,
    fonttitle=\bfseries,
    coltitle=black,
    fontupper=\footnotesize
]
\textbf{System prompt:}

You are a clinical pharmacist safety reviewer. Given a list of predicted
prescription drugs and safety signals (drug--drug interactions and
contraindications), decide which drugs to \textsc{remove} or \textsc{retain}.

\vspace{0.5em}
\textbf{Adjudication rules:}
\begin{itemize}[leftmargin=1.2em,itemsep=0.15em,topsep=0.2em]
    \item If a drug is in \texttt{prior\_meds} (the patient was on it before
    this admission), prefer to \textsc{keep} it unless the interaction is
    severe.

    \item When a DDI pair is flagged, prefer removing the drug with the
    higher global DDI degree (more interactions overall) and lower clinical
    necessity.

    \item When a drug is contraindicated with a patient diagnosis, remove it
    unless it was in \texttt{prior\_meds} and the benefit clearly outweighs
    the risk.

    \item Be conservative: remove drugs only when there is a clear safety
    signal. Do not remove a drug merely because it has a low DDI degree.

    \item Return strict JSON only, with no markdown fences and no commentary.
\end{itemize}

\vspace{0.5em}
\textbf{Input (dynamically assembled user prompt):}

The user prompt contains four sections, assembled at runtime from the
candidate set and the two safety matrices:
\begin{enumerate}[leftmargin=1.4em,itemsep=0.15em,topsep=0.2em]
    \item \textbf{Predicted drugs.} Each candidate listed as
    \texttt{<ATC-L4>: <class name>}, with \texttt{[PRIOR-MED]} when the drug
    was active in the patient's previous admission. Example:
    \texttt{B01AB: Heparin group [PRIOR-MED]}.

    \item \textbf{Prior medications.} All drugs active in the previous visit,
    used for continuation decisions.

    \item \textbf{DDI pairs detected.} Each flagged pair from the binary DDI
    matrix, annotated with both drugs' global DDI degree and a
    \texttt{[PRIOR]} tag where applicable. Example: \texttt{B01AB
    (degree=42) [PRIOR] $\leftrightarrow$ N02BA (degree=38)}.

    \item \textbf{Contraindication pairs detected.} Each flagged
    drug--diagnosis pair from the binary contraindication matrix. Example:
    \texttt{A02BB $\leftrightarrow$ diagnosis O80}.
\end{enumerate}
If a case raises no DDI or contraindication flags, the candidate set is
retained without invoking the verifier.

\vspace{0.5em}
\textbf{Output format:}

Return strict JSON with two fields: \texttt{kept\_drugs} (list of retained
ATC-L4 codes) and \texttt{removed\_drugs} (list of objects, each with
\texttt{code}, a short \texttt{reason}, and an optional ATC-L4
\texttt{replacement} from the same therapeutic subgroup).
\end{tcolorbox}
\caption{
Prompt template for the \ours{} \textsc{Verify} operator described in
Section~\ref{sec:safety}. The system prompt encodes the retain/remove
adjudication policy; the user prompt is dynamically assembled from
matrix-retrieved DDI and contraindication flags, drug-name lookups, and
prior-medication status. A \texttt{[PRIOR]} tag marks medications carried
over from a prior visit so the model can bias toward continuation.}
\label{fig:prompt_safety_verifier}
\end{figure*}

\subsection{Baseline Prompt Templates}
\label{app:baseline_prompts}

We provide abbreviated prompt templates used in our experiments. Figures~\ref{fig:prompt_direct}--\ref{fig:prompt_rareagents} show the baseline templates for direct prompting, the general-agent baseline, and the adapted RareAgents baseline.

\begin{figure*}[t]
\begin{tcolorbox}[
    enhanced,
    width=\textwidth,
    colback=gray!5,
    colframe=gray!60,
    title=Direct LLM Prompt,
    fonttitle=\bfseries,
    coltitle=black,
    fontupper=\small,
]
\small
\textbf{System prompt:}

You are an inpatient physician. For an ICU inpatient admission, predict the ATC
Level-4 medication classes that will appear on this patient's medication list
during the admission.

\vspace{0.5em}
\textbf{User prompt:}

\{patient\_text\}

\vspace{0.5em}
\textbf{Task instruction:}

Think step by step, then emit a single JSON object. ATC L4 codes are exactly
5 characters: letter + 2 digits + 2 letters (e.g., C10AA, B01AB, N02BE).
Do not invent codes, if unsure, omit.

\vspace{0.5em}
\textbf{Output format:}

Output STRICT JSON only (no markdown, no commentary):
\begin{verbatim}
{
  "reasoning": "<step-by-step reasoning, plain text>",
  "predicted_drugs": [
    {"code": "<ATC L4>"}
  ]
}
\end{verbatim}
\end{tcolorbox}
\caption{Prompt template for the direct prompting baseline.}
\label{fig:prompt_direct}
\end{figure*}

\begin{figure*}[t]
\begin{tcolorbox}[
    enhanced,
    width=\textwidth,
    colback=gray!5,
    colframe=gray!60,
    title=General-Agent Prompt,
    fonttitle=\bfseries,
    coltitle=black,
    fontupper=\small,
]
\small
\textbf{Note.}
The general-agent baseline reuses \ours{}'s \textsc{Summarize} and
\textsc{Generate} prompt structure
(Figures~\ref{fig:prompt_summarizer} and~\ref{fig:prompt_generator}) and the
shared \textsc{Critique} prompt (Figure~\ref{fig:prompt_critique}), but
replaces the routed specialty experts with a single \emph{general\_agent}
skill that covers every therapeutic domain. Only the skill's injected
\emph{summarization playbook} and \emph{drug-prediction checklist} differ
from the SafeRx-Agent versions; both are reproduced below.

\vspace{0.7em}
\textbf{\textsc{Summarize} stage --- general-agent playbook:}

Summarize the patient record to support broad inpatient medication prediction.
The summary should capture active conditions, relevant chronic history,
procedures, acuity of care, medication-relevant risks, and prior medication
evidence. It should be comprehensive across therapeutic domains rather than
specialized to one organ system.

In particular, the summary should preserve information that may affect
medication choice, including acute diagnoses, comorbidities, procedures,
critical-care status, infection evidence, pain or neurologic needs, renal and
metabolic abnormalities, gastrointestinal or nutritional status, VTE or bleeding
risk, and medications that may require continuation from prior visits.

\vspace{0.7em}
\textbf{\textsc{Generate} stage --- general-agent drug-prediction instruction:}

Predict ATC-L4 medication classes for the current inpatient admission as a
single generalist agent. Use the patient summary, prior-visit medication
evidence, diagnosis-linked medication evidence, and the closed ATC-L4
vocabulary. The prediction should cover both disease-directed therapy and
common inpatient supportive care, while requiring patient-specific evidence for
each medication class.

The agent should reason over the case by identifying active problems and
chronic conditions, mapping each to plausible medication needs, checking whether
prior medications should be continued, and considering medications associated
with procedures, critical illness, infection management, pain control,
thrombosis prevention, gastrointestinal protection, electrolyte correction,
nutrition, and bowel care. Medication classes should be included only when
supported by the current visit context, longitudinal history, or explicit
evidence in the record.

The agent should prefer adequate coverage of clinically supported medication
needs, but should avoid adding classes solely because they are common in
inpatient care. Confidence should reflect the strength of evidence: higher
confidence is reserved for explicitly documented medications or indications,
while lower confidence is used for plausible but indirect clinical support.
\end{tcolorbox}
\caption{Prompt template for the general-agent baseline. The baseline reuses
\ours{}'s \textsc{Summarize}, \textsc{Generate}, and \textsc{Critique}
structures, but replaces routed specialty experts with one all-domain
\emph{general\_agent}.}
\label{fig:prompt_general_agent}
\end{figure*}

\begin{figure*}[t]
\begin{tcolorbox}[
    enhanced,
    width=\textwidth,
    colback=gray!5,
    colframe=gray!60,
    title=RareAgents-Style Prompts Adapted to ATC-L4,
    fonttitle=\bfseries,
    coltitle=black,
    fontupper=\small,
]
\small

\textbf{Stage 1: Attending Physician.}\quad
You are the Attending Physician coordinating a multi-disciplinary team (MDT).
Given the patient profile and a pool of 41 clinical specialists, each defined by
an \texttt{id}, name, and clinical scope, select the specialists whose expertise
is relevant to the patient's active diagnoses, procedures, and medication needs.
Prefer disease- and organ-system specialists when they directly match the case,
and include cross-cutting specialists such as clinical pharmacy or internal
medicine when their scope can improve treatment review.
Return the selected specialist IDs and a brief rationale as JSON.

\vspace{0.5em}\noindent\rule{\linewidth}{0.3pt}\vspace{0.5em}

\textbf{Stage 2: Specialist Discussion.}\quad
You are the \texttt{\{specialist\_name\}} specialist in the MDT. Based on the
patient profile, prior prescription history, and your specialty scope, discuss
which medication classes are clinically relevant for the current admission.
Use the provided DrugBank and DDI-graph feedback, when available, to note
treatment evidence and potential safety concerns. Propose only medications
within your specialty scope. Output 5-character ATC-L4 codes adapted to our
closed prediction vocabulary, and convert ingredient-level ATC codes to their
L4 parent. Return an empty list if your specialty has no relevant role. Return
the proposed codes and reasoning as JSON.

\vspace{0.5em}\noindent\rule{\linewidth}{0.3pt}\vspace{0.5em}

\textbf{Stage 3: Attending Synthesis.}\quad
You are the Attending Physician finalizing the medication recommendation after
the MDT discussion. Integrate the specialist proposals, prior prescription
evidence, and DrugBank/DDI-graph feedback to form the final prescription set.
Use the specialist proposals as the candidate pool, keep medications supported
by the patient context or prior-visit precedent, and remove unsupported or
tool-flagged candidates when the evidence does not justify keeping them.
Do not introduce medications that were not proposed by any specialist.
Output 1--30 ATC-L4 codes with reasoning as JSON.

\end{tcolorbox}
\caption{
\textbf{RareAgents-style baseline prompts adapted to ATC-L4 medication
recommendation.}
We adapt the RareAgents MDT workflow~\cite{rareagents} to our closed-vocabulary
ATC-L4 setting. The three stages correspond to attending-led specialist
selection, specialist discussion with DrugBank and DDI-graph feedback, and
attending-led synthesis. Unlike the original RareAgents medication prompt, which
selects medication names from a fixed candidate list, our adaptation requires
5-character ATC-L4 codes from the prediction vocabulary.
}
\label{fig:prompt_rareagents}
\end{figure*}

\section{Case Study}
\label{app:case_analysis}

We present an end-to-end walkthrough of \ours{} on a representative MIMIC-IV case (Subject~10785159) using Gemma3-27B-IT. This case illustrates how multi-expert routing, grounded generation, and global critique produce a traceable medication set with per-medication rationales. Figure~\ref{fig:case_pipeline} provides an overview of the pipeline flow on this case; Table~\ref{tab:case_proposals} traces every candidate code through the system.

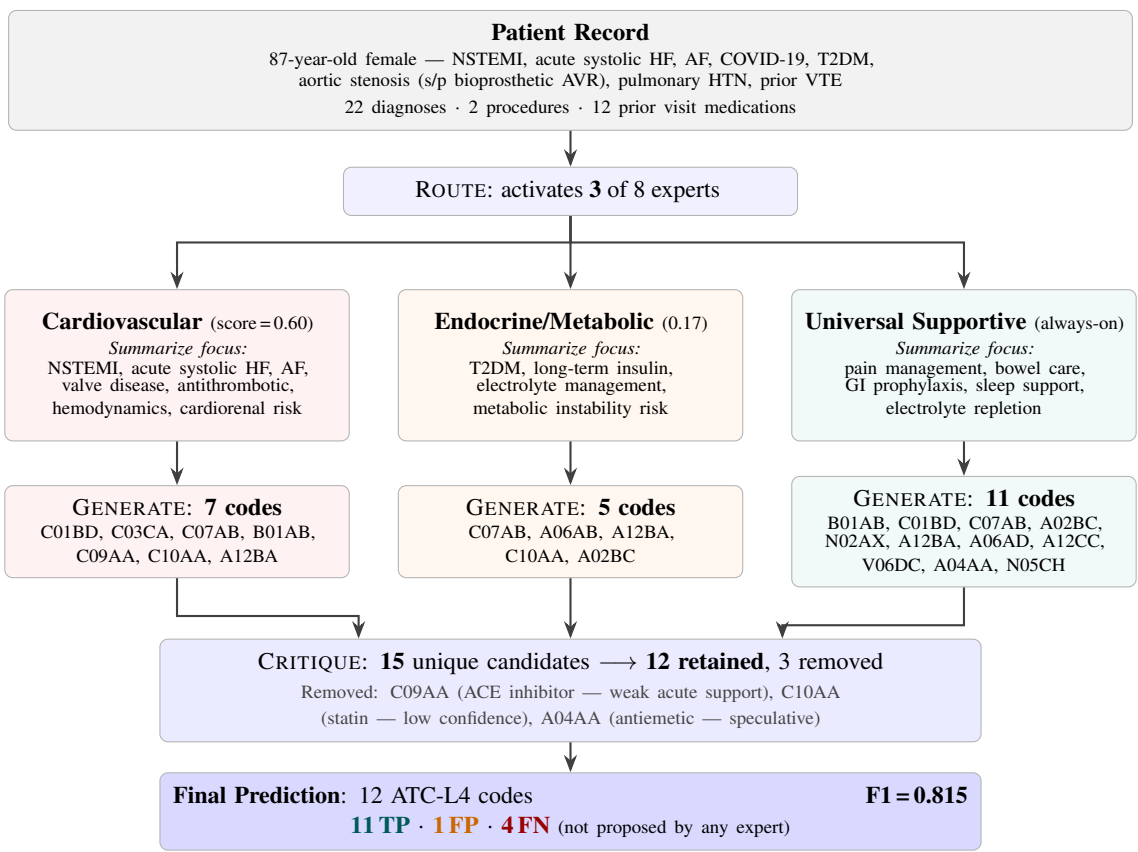
\begin{figure*}[t]
\centering
\begin{tikzpicture}[
    >=Stealth,
    box/.style={rectangle, draw=gray!60, rounded corners=3pt, inner sep=5pt, font=\small, align=center},
    arr/.style={->, thick, gray!50!black}
]

% Row 1: Patient
\node[box, fill=gray!10, text width=14.5cm] (P) at (0,0) {
    \textbf{Patient Record}\\[0.15em]
    \scriptsize 87-year-old female --- NSTEMI, acute systolic HF, AF, COVID-19, T2DM, aortic stenosis (s/p bioprosthetic AVR), pulmonary HTN, prior VTE\\
    \scriptsize 22 diagnoses $\cdot$ 2 procedures $\cdot$ 12 prior visit medications
};

% Row 2: Route
\node[box, fill=blue!6, minimum width=6cm] (R) at (0,-1.6) {
    \textsc{Route}: activates \textbf{3} of 8 experts
};

% Row 3: Expert + Summarize
\node[box, fill=red!5, text width=4.2cm, minimum height=2cm] (CV) at (-5.2,-3.9) {
    \textbf{Cardiovascular} \scriptsize(score\,=\,0.60)\\[0.2em]
    \scriptsize\textit{Summarize focus:}\\[-0.1em]
    \scriptsize NSTEMI, acute systolic HF, AF,\\[-0.2em]
    \scriptsize valve disease, antithrombotic,\\[-0.2em]
    \scriptsize hemodynamics, cardiorenal risk
};
\node[box, fill=orange!6, text width=4.2cm, minimum height=2cm] (EN) at (0,-3.9) {
    \textbf{Endocrine/Metabolic} \scriptsize(0.17)\\[0.2em]
    \scriptsize\textit{Summarize focus:}\\[-0.1em]
    \scriptsize T2DM, long-term insulin,\\[-0.2em]
    \scriptsize electrolyte management,\\[-0.2em]
    \scriptsize metabolic instability risk
};
\node[box, fill=teal!5, text width=4.2cm, minimum height=2cm] (UN) at (5.2,-3.9) {
    \textbf{Universal Supportive} \scriptsize(always-on)\\[0.2em]
    \scriptsize\textit{Summarize focus:}\\[-0.1em]
    \scriptsize pain management, bowel care,\\[-0.2em]
    \scriptsize GI prophylaxis, sleep support,\\[-0.2em]
    \scriptsize electrolyte repletion
};

% Row 4: Generate
\node[box, fill=red!5, text width=4.2cm] (CVg) at (-5.2,-6.1) {
    \textsc{Generate}: \textbf{7 codes}\\[0.05em]
    \scriptsize C01BD, C03CA, C07AB, B01AB,\\[-0.15em]
    \scriptsize C09AA, C10AA, A12BA
};
\node[box, fill=orange!6, text width=4.2cm] (ENg) at (0,-6.1) {
    \textsc{Generate}: \textbf{5 codes}\\[0.05em]
    \scriptsize C07AB, A06AB, A12BA,\\[-0.15em]
    \scriptsize C10AA, A02BC
};
\node[box, fill=teal!5, text width=4.2cm] (UNg) at (5.2,-6.1) {
    \textsc{Generate}: \textbf{11 codes}\\[0.05em]
    \scriptsize B01AB, C01BD, C07AB, A02BC,\\[-0.15em]
    \scriptsize N02AX, A12BA, A06AD, A12CC,\\[-0.15em]
    \scriptsize V06DC, A04AA, N05CH
};

% Row 5: Critique
\node[box, fill=blue!8, text width=10.5cm, minimum height=1.1cm] (C) at (0,-8.2) {
    \textsc{Critique}: \textbf{15} unique candidates $\longrightarrow$ \textbf{12 retained}, 3 removed\\[0.1em]
    \scriptsize\textcolor{gray!55!black}{Removed: C09AA (ACE inhibitor --- weak acute support), C10AA (statin --- low confidence), A04AA (antiemetic --- speculative)}
};

% Row 6: Final
\node[box, fill=blue!15, text width=10.5cm] (F) at (0,-9.8) {
    \textbf{Final Prediction}: 12 ATC-L4 codes\hfill\textbf{F1\,=\,0.815}\\[0.15em]
    \small \textcolor{teal!70!black}{\textbf{11\,TP}} $\cdot$
    \textcolor{orange!80!black}{\textbf{1\,FP}} $\cdot$
    \textcolor{red!60!black}{\textbf{4\,FN}} \scriptsize(not proposed by any expert)
};

% Arrows: Patient -> Route
\draw[arr] (P) -- (R);
% Route -> Experts (fan-out)
\draw[arr] (R.south) -- ++(0,-0.35) -| (CV.north);
\draw[arr] (R.south) -- (EN.north);
\draw[arr] (R.south) -- ++(0,-0.35) -| (UN.north);
% Experts -> Generate
\draw[arr] (CV) -- (CVg);
\draw[arr] (EN) -- (ENg);
\draw[arr] (UN) -- (UNg);
% Generate -> Critique (fan-in)
\draw[arr] (CVg.south) -- ++(0,-0.5) -| ([xshift=-2.8cm]C.north);
\draw[arr] (ENg.south) -- (C.north);
\draw[arr] (UNg.south) -- ++(0,-0.5) -| ([xshift=2.8cm]C.north);
% Critique -> Final
\draw[arr] (C) -- (F);

\end{tikzpicture}
\caption{\textbf{Case study pipeline flow.} The patient record is routed to three specialty experts. Each expert summarizes the record from its domain perspective and generates ATC-L4 candidates. \textsc{Critique} merges the 15 unique candidates across all experts and removes 3 weakly supported codes, producing a final set of 12 medications.}
\label{fig:case_pipeline}
\end{figure*}

\subsection{Patient Presentation}

\begin{tcolorbox}[
    enhanced,
    colback=gray!5,
    colframe=gray!50,
    title={\includegraphics[height=1.4em]{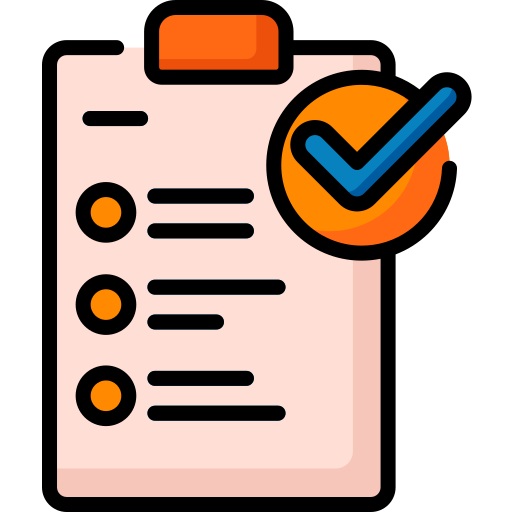}\ Patient Vignette},
    fonttitle=\bfseries\small
]
\small
An 87-year-old female with a history of nonrheumatic aortic stenosis (post-bioprosthetic valve replacement), chronic heart failure, persistent atrial fibrillation, cardiomyopathy, hypertensive heart disease, type~2 diabetes mellitus on insulin, hyperlipidemia, obesity, GERD, irritable bowel syndrome, and prior venous thromboembolism. She presents with \textbf{NSTEMI}, acute-on-chronic systolic heart failure, COVID-19, pulmonary hypertension, phlebitis of the lower extremities, and hypotension. Current visit procedures include coronary angiography and cardiac catheterization. Prior visit medications (12 codes) include amiodarone (C01BD), furosemide (C03CA), metoprolol (C07AB), heparin (B01AB), PPI (A02BC), potassium (A12BA), magnesium (A12CC), tramadol (N02AX), contact laxatives (A06AB), osmotic laxatives (A06AD), melatonin (N05CH), and carbohydrate supplement (V06DC).

\vspace{0.3em}
\textbf{Ground-truth medications for the current visit}: 15 ATC-L4 codes.
\end{tcolorbox}

\subsection{Pipeline Walkthrough}

\paragraph{Step~1: Expert Routing (\textsc{Route}).}
\textsc{Route} scores the patient's ICD codes against each expert's scope. The cardiovascular domain dominates (score\,=\,0.60), followed by endocrine/metabolic (0.17). Three experts are activated: \textit{Cardiovascular}, \textit{Endocrine/Metabolic}, and the always-on \textit{Universal Supportive Care} (Figure~\ref{fig:case_pipeline}, rows~2--3).

\paragraph{Step~2: Expert Summarization (\textsc{Summarize}).}
Each activated expert extracts specialty-relevant evidence from the patient record:

\begin{itemize}[leftmargin=1.5em, itemsep=0.1em, topsep=0.2em]
    \item \textbf{Cardiovascular} identifies NSTEMI, acute systolic HF, AF, aortic stenosis, pulmonary HTN, and hypotension as active problems. It notes the prior aortic valve replacement and long-term anticoagulant use, and flags risks of bleeding (anticoagulation plus recent procedure), arrhythmia (AF, post-MI), and cardiorenal syndrome.
    \item \textbf{Endocrine/Metabolic} focuses on type~2 diabetes with long-term insulin use, and flags electrolyte management (prior potassium and magnesium supplementation) and metabolic instability risks given acute heart failure.
    \item \textbf{Universal Supportive Care} provides a broad view covering pain management (NSTEMI), bowel care (IBS history plus opioid use), sleep support, GI prophylaxis (GERD plus anticoagulation), and electrolyte repletion in the context of heart failure and anticipated diuretic use.
\end{itemize}

\paragraph{Step~3: Per-Expert Generation (\textsc{Generate}).}
Each expert proposes ATC-L4 codes grounded in its summary and the MEDI indication resource. The Cardiovascular expert proposes 7 codes, the Endocrine/Metabolic expert proposes 5 codes, and the Universal Supportive Care expert proposes 11 codes, yielding 15 unique candidates across all experts (Table~\ref{tab:case_proposals}).

\paragraph{Step~4: Global Critique (\textsc{Critique}).}
\textsc{Critique} reviews all expert proposals under the full patient context and removes three candidates:
\begin{itemize}[leftmargin=1.5em, itemsep=0.1em, topsep=0.2em]
    \item \textbf{C09AA} (ACE inhibitors): while plausible given diabetes and cardiovascular disease, the acute presentation (hypotension, acute HF) does not strongly support initiation, and no prior ACE inhibitor use is documented.
    \item \textbf{C10AA} (statins): both proposing experts assigned low confidence, and the acute presentation does not prioritize statin initiation.
    \item \textbf{A04AA} (antiemetics): a speculative proposal, as COVID-19 alone does not necessitate antiemetics without documented nausea or vomiting.
\end{itemize}
The final candidate set contains 12 ATC-L4 codes.

\begin{table*}[t]
    \centering
    \small
    \setlength{\tabcolsep}{4pt}
    \caption{\textbf{Case study pipeline trace.} Checkmarks indicate which expert(s) proposed each ATC-L4 code via \textsc{Generate}. \textsc{Critique} action shows whether the code was retained (\textsc{Ret}) or removed (\textsc{Rem}). Outcome classifies each code against the ground-truth medication set: true positive (TP), false positive (FP), or not applicable (---) for correctly removed candidates. False-negative medications (bottom) were not proposed by any expert.}
    \label{tab:case_proposals}
    \begin{tabular}{llcccccl}
    \toprule
    \textbf{ATC-L4} & \textbf{Drug Class} & \textbf{CV} & \textbf{Endo} & \textbf{Univ} & \textbf{\#Exp} & \textbf{Critique} & \textbf{Outcome} \\
    \midrule
    C07AB & Beta-blockers, selective & \checkmark & \checkmark & \checkmark & 3 & \textsc{Ret} & TP \\
    A12BA & Potassium & \checkmark & \checkmark & \checkmark & 3 & \textsc{Ret} & TP \\
    B01AB & Heparin group & \checkmark & & \checkmark & 2 & \textsc{Ret} & TP \\
    C01BD & Antiarrhythmics, class~III & \checkmark & & \checkmark & 2 & \textsc{Ret} & TP \\
    A02BC & Proton pump inhibitors & & \checkmark & \checkmark & 2 & \textsc{Ret} & TP \\
    C10AA & HMG CoA reductase inhibitors & \checkmark & \checkmark & & 2 & \textsc{Rem} & --- \\
    \midrule
    C03CA & Sulfonamides, plain (loop diuretics) & \checkmark & & & 1 & \textsc{Ret} & TP \\
    C09AA & ACE inhibitors, plain & \checkmark & & & 1 & \textsc{Rem} & --- \\
    A06AB & Contact laxatives & & \checkmark & & 1 & \textsc{Ret} & FP \\
    N02AX & Other opioids & & & \checkmark & 1 & \textsc{Ret} & TP \\
    A06AD & Osmotically acting laxatives & & & \checkmark & 1 & \textsc{Ret} & TP \\
    A12CC & Magnesium & & & \checkmark & 1 & \textsc{Ret} & TP \\
    V06DC & Carbohydrates & & & \checkmark & 1 & \textsc{Ret} & TP \\
    N05CH & Melatonin receptor agonists & & & \checkmark & 1 & \textsc{Ret} & TP \\
    A04AA & Serotonin (5-HT\textsubscript{3}) antagonists & & & \checkmark & 1 & \textsc{Rem} & --- \\
    \midrule
    \multicolumn{8}{l}{\textit{Ground-truth medications not proposed by any expert (false negatives):}} \\
    A06AX & Other drugs for constipation & & & & 0 & --- & FN \\
    D01AC & Imidazole/triazole antifungals & & & & 0 & --- & FN \\
    J07BX & Other viral vaccines & & & & 0 & --- & FN \\
    N06AX & Other antidepressants & & & & 0 & --- & FN \\
    \bottomrule
    \end{tabular}
\end{table*}

\subsection{Analysis}

The final predicted set contains 12 medications, achieving a precision of 0.917, recall of 0.733, and F1 of 0.815. Figure~\ref{fig:case_support} summarizes the relationship between expert support and prediction outcome.

\paragraph{Multi-expert corroboration.}
All five codes proposed by $\geq$2 experts that were retained by \textsc{Critique} are true positives (C07AB, A12BA, B01AB, C01BD, A02BC), consistent with the aggregate finding that multi-expert corroboration is a strong signal for correctness (Section~\ref{sec:diagnostic_analysis}).

\paragraph{Selective critique.}
\textsc{Critique} removes 3 of the 15 candidate codes. None of the three removed drugs appear in the ground truth, indicating that the \textsc{Critique} correctly identified false positives in this case. Among single-expert proposals, \textsc{Critique} retains those with clear patient-specific support---for example, C03CA (loop diuretics) was proposed only by the Cardiovascular expert but was kept because the patient has documented heart failure, volume overload indicators, and prior furosemide use. In contrast, C09AA (ACE inhibitors) was removed despite cardiovascular relevance because the patient's hypotension and lack of prior ACE inhibitor use argued against initiation.

\paragraph{False positive.}
The single false positive, A06AB (contact laxatives), was prescribed in the prior visit and proposed by the Endocrine/Metabolic expert based on documented constipation history. This is a clinically reasonable continuation that falls outside the current visit's ground-truth set, illustrating that not all false positives represent clinical errors.

\paragraph{False negatives.}
The four missed medications---A06AX (other drugs for constipation), D01AC (imidazole/triazole antifungals), J07BX (other viral vaccines), and N06AX (other antidepressants)---were not proposed by any expert. D01AC and J07BX lack explicit diagnostic indicators in the structured record (antifungal prophylaxis and COVID-19 vaccination are context-dependent decisions). N06AX (antidepressants) may reflect undocumented psychiatric history. These false negatives highlight the challenge of predicting medications whose indications require implicit clinical reasoning beyond the available EHR data.

\end{document}